\documentclass[journal,twoside,web]{ieeecolor}

\usepackage{jsen}
\usepackage{cite}
\usepackage{amsmath,amssymb,amsfonts}
\usepackage{algorithmic}
\usepackage{graphicx}
\usepackage[outdir=./]{epstopdf}

\usepackage{textcomp}
\usepackage{wrapfig}
\usepackage{multirow}
\usepackage{subfigure}
\usepackage{url, footnote}
\usepackage{bm}
\usepackage{makecell}
\def\BibTeX{{\rm B\kern-.05em{\sc i\kern-.025em b}\kern-.08em
    T\kern-.1667em\lower.7ex\hbox{E}\kern-.125emX}}
\definecolor{abstractbg}{rgb}{0.89804,0.94510,0.83137}
\begin{document}
\title{TransFusionOdom: Interpretable Transformer-based LiDAR-Inertial Fusion Odometry Estimation}  
\author{Leyuan Sun, Guanqun Ding, Yue Qiu, Yusuke Yoshiyasu and Fumio Kanehiro
\thanks{This research was partially supported by a research project grant from the JST-SPRING, the grant number is JPMJSP2124. (\textit{Corresponding author: Leyuan Sun}) }
\thanks{Leyuan Sun and Fumio Kanehiro are with CNRS-AIST Joint Robotics Laboratory (JRL), IRL, National Institute of Advanced Industrial Science and Technology (AIST), Tsukuba, Japan (e-mail: son.leyuansun, f-kanehiro@aist.go.jp)}
\thanks{Leyuan Sun, Yue Qiu and Yusuke Yoshiyasu are with the Computer Vision Research Team, Artificial Intelligence Research Center (AIRC), National Institute of Advanced Industrial Science and Technology (AIST), Tsukuba, Japan (e-mail: qiu.yue, yusuke-yoshiyasu@aist.go.jp)}
\thanks{Guanqun Ding is with the Digital Architecture Research Center (DigiARC), National Institute of Advanced Industrial Science and Technology (AIST), Tokyo, Japan. (e-mail: guanqun.ding@aist.go.jp)}
\thanks{Fumio Kanehiro is with Department of Intelligent and Mechanical Interaction Systems, Graduate School of Science and Technology, University of Tsukuba, Tsukuba, Ibaraki, Japan}
}

\IEEEtitleabstractindextext{%
\fcolorbox{abstractbg}{abstractbg}{%
\begin{minipage}{\textwidth}%
\begin{wrapfigure}[11]{r}{3.6in}%
\includegraphics[width=3.5in]{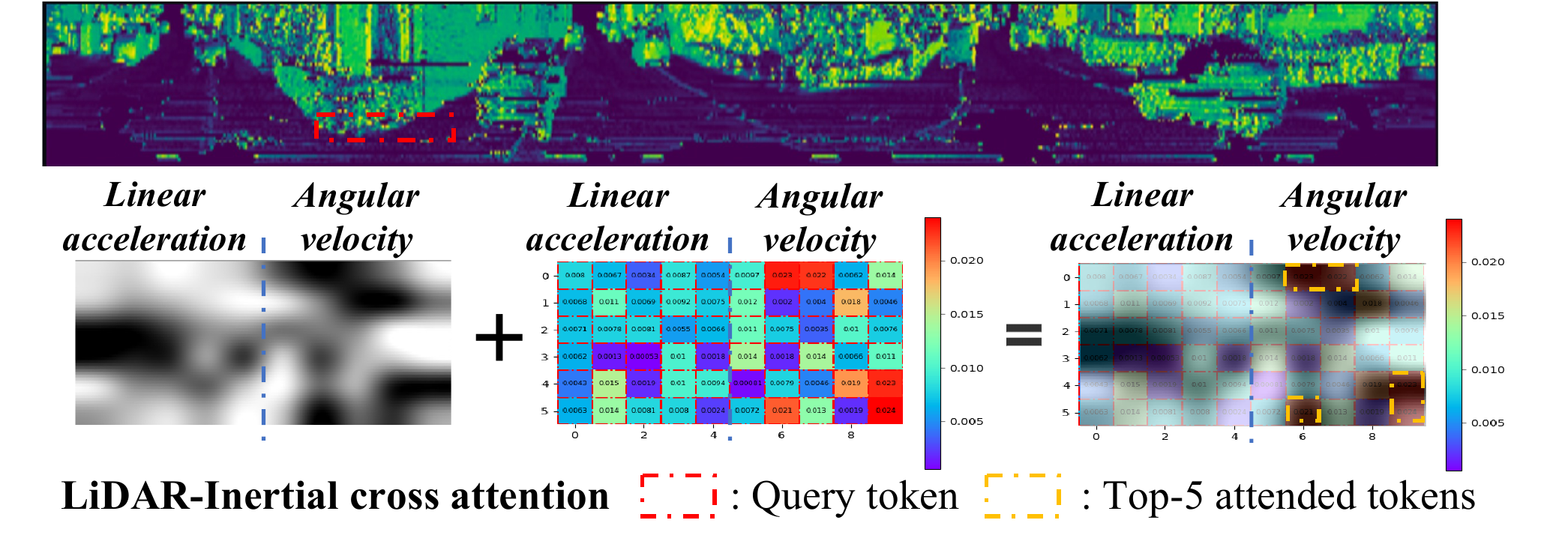}%
\end{wrapfigure}%
\begin{abstract}
Multi-modal fusion of sensors is a commonly used approach to enhance the performance of odometry estimation, which is also a fundamental module for mobile robots. However, the question of \textit{how to perform fusion among different modalities in a supervised sensor fusion odometry estimation task?} is still one of challenging issues remains. Some simple operations, such as element-wise summation and concatenation, are not capable of assigning adaptive attentional weights to incorporate different modalities efficiently, which make it difficult to achieve competitive odometry results. Recently, the Transformer architecture has shown potential for multi-modal fusion tasks, particularly in the domains of vision with language. In this work, we propose an end-to-end supervised Transformer-based LiDAR-Inertial fusion framework (namely TransFusionOdom) for odometry estimation. The multi-attention fusion module demonstrates different fusion approaches for homogeneous and heterogeneous modalities to address the overfitting problem that can arise from blindly increasing the complexity of the model. Additionally, to interpret the learning process of the Transformer-based multi-modal interactions, a general visualization approach is introduced to illustrate the interactions between modalities. Moreover, exhaustive ablation studies evaluate different multi-modal fusion strategies to verify the performance of the proposed fusion strategy. A synthetic multi-modal dataset is made public to validate the generalization ability of the proposed fusion strategy, which also works for other combinations of different modalities. The quantitative and qualitative odometry evaluations on the KITTI dataset verify the proposed TransFusionOdom could achieve superior performance compared with other related works.

\end{abstract}

\end{minipage}}}

\maketitle

\section{Introduction}
\label{sec:introduction}
\IEEEPARstart{S}{ensor} fusion is a popular topic in robotics for several decades. In robotics, two types of sensors are commonly used, which are exteroceptive sensors and proprioceptive sensors. Exteroceptive sensors, such as cameras, LiDAR, ultrasonic sensors, and radar, provide rich and surrounding-sensitive information, but with a low frequency of around 30Hz. On the other hand, proprioceptive sensors, including Inertial Measurement Units (IMU), wheel odometers, and joint encoders, can estimate their own state at a high frequency of around 200Hz, but they tend to drift over time. Hence, combining these two types of sensors is widely implemented for specific tasks in robotics. In this paper, a sensor-fusion approach is proposed to enhance the performance of odometry estimation task.
\begin{figure}[!t]
	\centerline{\includegraphics[width=\columnwidth]{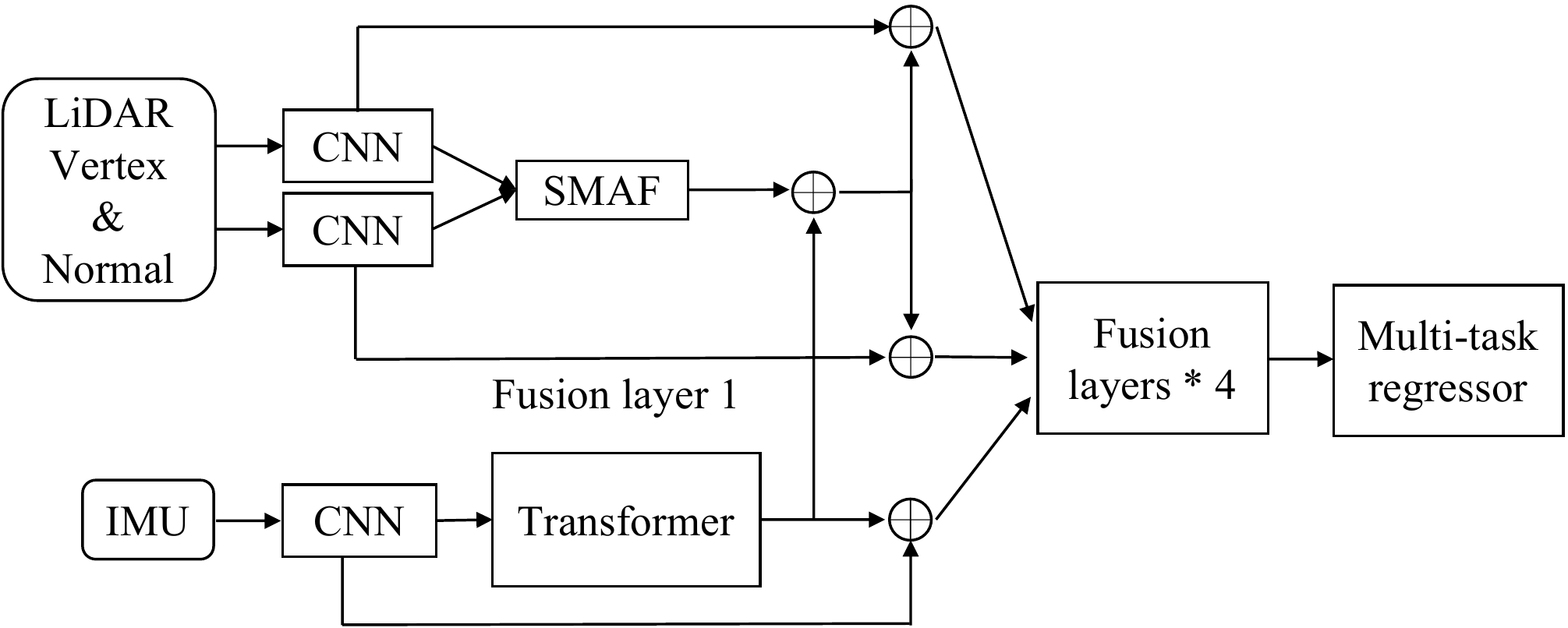}}
	\caption{A simple overview of proposed TransFusionOdom, input are LiDAR raw point cloud and IMU measurements, output are translation, orientation and their uncertainty estimations. }
	\label{overview}
\end{figure}

\begin{table*}[!t]
\caption{Related works on supervised sensor fusion for odometry estimation}
\label{relatedworks}
\renewcommand\arraystretch{1.5}
\begin{center}
\scalebox{0.85}{
\begin{tabular}{c|ccccc}
\hline
Supervised fusion odometry estimation                                            & \multicolumn{1}{c|}{Modalities} & \multicolumn{1}{c|}{Fusion strategy} & \multicolumn{1}{c|}{Fusion position} & \multicolumn{1}{c|}{Representation of rotation} & Distance function           \\ \hline
TransFusionOdom (ours)                                                                  & LiDAR+IMU                       & SMAF + Transformer-based             & Multi-layer fusion                   & Euler angle + se(3)                                   & L2                          \\ \cline{1-1}
EMA-VIO \cite{emavio}                                           & RGB+IMU                         & Transformer-based                    & Middle fusion                        & Euler angle                                     & L2                          \\ \cline{1-1}
\ AFT-VO \cite{aftvo}                                       & Multi RGBs                      & Transformer-based                    & Middle fusion                        & Not explain                                     & L2                          \\ \cline{1-1}
\ Chen \emph{et al.} \cite{selectivesensor} & RGB+IMU                         & Soft and Hard mask                   & Middle fusion                        & Quaternion                                      & L1                          \\ \cline{1-1}
Son \emph{et al.} \cite{sonsynthetic}    & LiDAR+IMU                       & Soft mask                            & Late fusion                          & Axis-angle                                      & L2                          \\ \cline{1-1}
DeepLIO \cite{deeplio}                                     & LiDAR+IMU                       & Soft mask                            & Middle fusion                        & Quaternion + se(3)                              & L2                          \\ \cline{1-1}
ATVIO \cite{atvio}                                        & RGB+IMU                         & Cross-domain attention mask          & Middle fusion                        & Euler angle                                     & Others                      \\ \cline{1-1}
VINet \cite{vinet}                                        & RGB+IMU                         & Concat.                              & Middle fusion                        & Quaternion + se(3)                              & L1                          \\ \cline{1-1}
Li \emph{et al.} \cite{li2019deep}        & Laser scan + IMU                & Concat.                              & Middle fusion                        & 1D angle                                        & L2                          \\ \cline{1-1}
VIIONet \cite{viionet}                                     & RGB+IMU image                   & Concat.                              & Middle fusion                        & Not explain                                     & Gaussian Process Regression \\ \cline{1-1}
HVIOnet \cite{hvionet}                                     & RGB+IMU                         & Concat.                              & Early fusion                         & Not explain                                     & L2                          \\ \hline
\end{tabular}
}
\end{center}
\end{table*}

Filter-based and optimization-based solutions have been traditionally employed for sensor fusion in robotics. However, the linearization errors in filter-based methods limit the accuracy of estimation \cite{viionet}. For optimization-based solutions, it is difficult to design adaptive weights, such as a covariance matrix, to balance each residual of the sensor \cite{d3vo}. Recently, numerous studies have confirmed the superiority of data-driven learning-based solutions over traditional solutions in odometry estimation, such as those proposed by \cite{emavio}, \cite{vinet}, \cite{deeplio}, and \cite{selectfusion}.

Although learning-based frameworks offer several advantages such as non-calibration and initialization, robust feature extraction, and reduced time consumption, \textit{How should we perform fusion among different modalities in a supervised sensor fusion odometry estimation task?} is still one of the key issues remains, which we introduce and comprehensively evaluate in this study.

In this work, we focus on the fusion module in the learning-based multi-modal fusion field, which is also a general component of tasks that uses multiple modalities as input. In contrast to operate simple element-wise summation and concatenation, the proposed fusion strategy generates the adaptive weights or attentions between different modalities. The overview of the proposed TransFusionOdom is shown in Fig. \ref{overview}. To achieve homogeneous data fusion between LiDAR's vertex and normal estimation, we implement the Soft Mask Attentional Fusion (SMAF) \cite{selectivesensor} \cite{pointloc} \cite{selfattfuse}. For heterogeneous data fusion between LiDAR and IMU, we introduce the Transformer \cite{transformer} encoder architecture. This multi-attention fusion approach is designed to avoid the overfitting problem, which is prone to occur if we overly increase the complexity of Transformer-based fusion network \cite{dropkey}.

The main contributions of this article can be summarized as follows:

\begin{enumerate}
  \item An end-to-end Transformer-based supervised multi-modal fusion odometry estimation network (TransFusionOdom) is developed, which achieves the competitive or even superior performance on KITTI dataset. The multi-attention fusion module is proposed to achieve the adaptive weights learning for the fusion between homogeneous and heterogeneous modalities. 
  \item A general visualization approach is demonstrated the interactions between two modalities inside the Transformer architecture, which enhances the interpretability of the proposed learning-based solution.
  \item A synthetic multi-modal dataset is publicly available \footnote{\url{https://github.com/RakugenSon/Multi-modal-dataset-for-odometry-estimation}}, which is used to evaluate the generalization ability of the proposed fusion strategy with different combinations of modalities. This dataset also enables easy testing of other fusion algorithms within the community.
  \item Exhaustive ablation studies are conducted on each module of TransFusionOdom, as well as on different representations of rotation and distance functions in the supervised sensor fusion odometry estimation task. 
\end{enumerate}

\begin{figure*}[!t]
	\centerline{\includegraphics[width=1\textwidth]{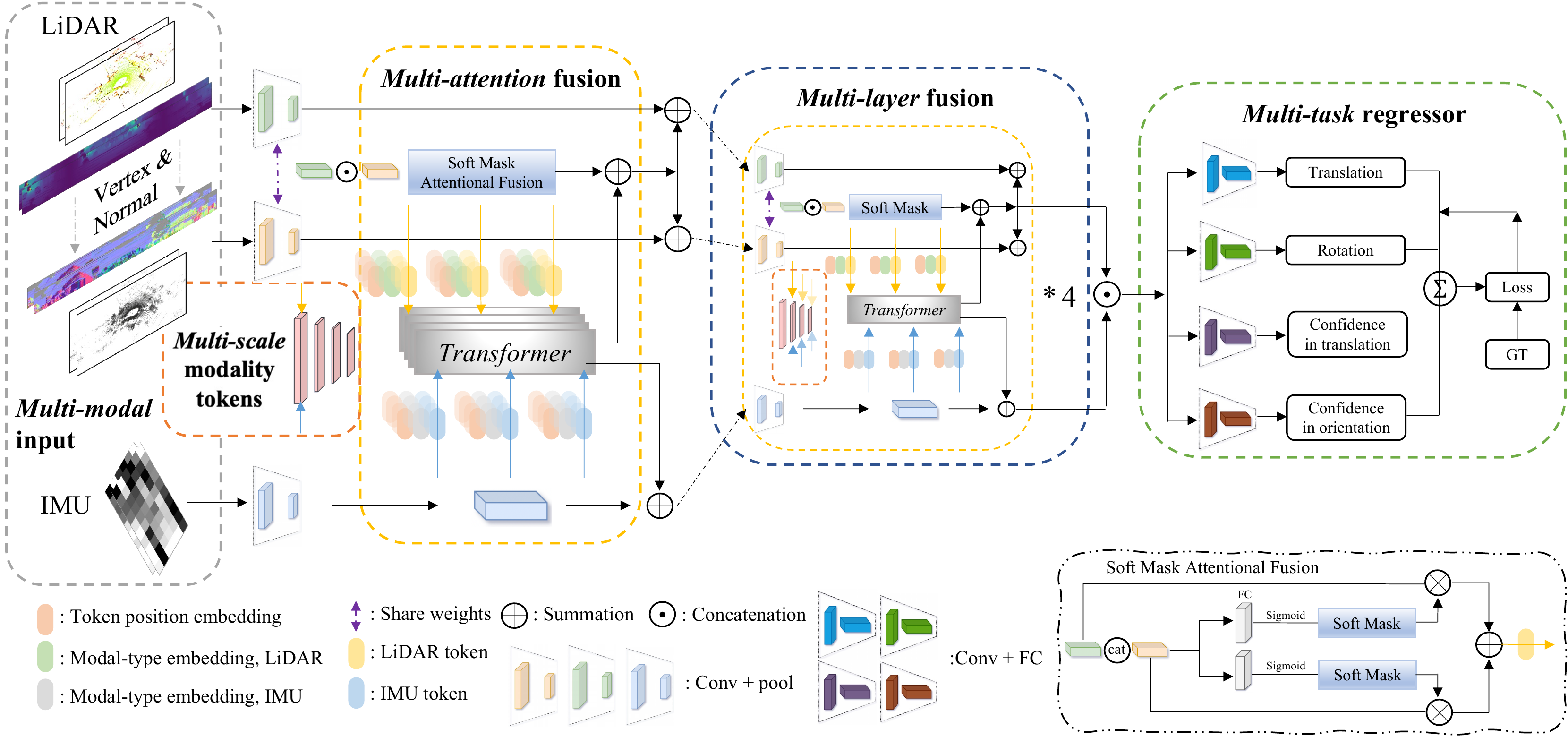}}
	\caption{Network architecture of proposed TransFusionOdom.}
	\label{architecture}
\end{figure*}

\section{Related work}
In this section, we provide a brief introduction to geometry-based and learning-based odometry estimation. Since our work primarily focuses on sensor fusion, we discuss traditional sensor fusion approaches in robotics, as well as the integration of existing learning-based supervised multi-modal fusion for odometry estimation, which is presented in Table \ref{relatedworks}.

\subsection{Geometry-based and learning-based odometry estimation}

In multi-view geometry-based odometry estimation, such as ORB-SLAM2 \cite{orb2} and RTAB-Map \cite{rtab}, the front-end consists of sensor calibration, keypoints extraction and matching, and outlier rejection modules. Under ideal conditions, where the environment is static \cite{SII2020} \cite{jrm} and there are no issues with texture or illumination, the accuracy of odometry estimation is satisfactory \cite{viionet}. However, real-world applications require high robustness of the odometry estimation system to deal with challenging scenarios. Recently, data-driven learning-based approaches have integrated all front-end modules into an end-to-end model. Moreover, starting from DeepVO \cite{deepvo}, CNN-based feature extraction backbones are capable to learn the latent representation of input consecutive frames, which are more robust than manually designed features such as ORB, SIFT, and SURF. Additionally, in unsupervised SC-Depth \cite{scdepth}, the photometric loss can even be used to mask dynamic objects, making the framework robust to robot-and-human interactive environments.

\subsection{Traditional sensor fusion in robotics}
As mobile robot platforms are being equipped with an increasing number of sensors, researchers in robotics are recognizing the complementary abilities of certain sensors to improve the accuracy and robustness of estimation. One popular sensor combination is between vision-based sensors and IMUs, as can be observed in systems like VINS \cite{vins}, ORB-SLAM3 \cite{orbslam3}, LIO-SAM \cite{liosam}, and LeGO-LOAM \cite{legoloam}.

Traditional filter-based sensor fusion approaches, such as Kalman Filter and even more advanced MSCKF \cite{msckf}, suffer from linearization errors and computation cost problems. In recent years, optimization-based solutions have become the mainstream, as seen in systems such as VINS \cite{vins}, ORB-SLAM3 \cite{orbslam3}, and LIO-SAM \cite{liosam}. One limitation of optimization-based methods is that the weight of each sensor is difficult to be adaptive in real-time, making it challenging to represent the reliability of each sensor. To address this issue, D3VO \cite{d3vo} utilize the uncertainties obtained from a learning-based solution as the weights of each sensor's residual in the back-end optimization. However, to achieve this, the system is separated into two discrete stages: uncertainty learning and optimization, instead of an end-to-end manner.

\subsection{Learning-based multi-modal fusion for odometry estimation}
In Table \ref{relatedworks}, we list some related works on supervised sensor fusion odometry estimation. Since this article mainly focuses on the effect of fusion in this task, these related works have been categorized based on their fusion strategies and positions.

\subsubsection{How to fuse multiple modalities}
Under the category of fusion strategies, there are concatenate-based approaches such as VINet\cite{vinet}, Li \emph{et al.}'s method \cite{li2019deep}, VIIONet\cite{viionet}, and HVIOnet \cite{hvionet}. These methods directly concatenate the features from different modalities, which means that the weight of each sensor is constant and equal.

On the other hand, Chen \emph{et al.} propose \cite{selectivesensor}, Son \emph{et al.} develop \cite{sonsynthetic}, DeepLIO \cite{deeplio}, and ATVIO \cite{atvio} use the soft mask-based method, which relies on multilayer perceptron (MLP) to learn the weights of each sensor. These methods are similar to self-attention in PointLoc \cite{pointloc}. The learnable mask could remove outliers by giving low weight, and the adaptive reliability of each sensor during fusion improves the accuracy and robustness of odometry estimation through attentional mechanisms.

However, the ability of the simple MLP-based attentional mask architecture is not sufficient to handle challenging situations in the real world, such as reflection ground and overcast days \cite{emavio}. Inspired by ViLT \cite{vilt}, the Transformer \cite{transformer} architecture has shown impressive performance in the field of multi-modal fusion, not limited to odometry estimation tasks but also in navigation \cite{fukushima2022object}, semantic segmentation, and object detection tasks \cite{wang2022multimodal}. In EMA-VIO \cite{emavio} and AFT-VO \cite{aftvo}, the Transformer architecture is used to fuse multiple modalities, and through challenging real-world experiments, it has shown higher accuracy and robustness than some soft mask-based approaches. But these works did not consider the effect of fusion position, and the Transformer is used as a black box without interpretability to explain how two modalities interact and fusion inside the Transformer architecture.

\subsubsection{Where to conduct fusion}
Another important issue that needs to be taken into consideration is the fusion position. In Table \ref{relatedworks}, early fusion means fusing source modalities and then feeding them into the backbone for feature extraction. Middle fusion means using different backbones and fusing them before feeding into one regressor. If we feed into different regressors and then fuse them, it is defined as late fusion. Channel Exchange \cite{channelexchange} and MLF-VO \cite{mlfvo} have confirmed that multi-layer fusion is better than the previous three fusion positions. However, since the Transformer is a data-hungry model \cite{datahungry} compared to CNN-based models, implementing the Transformer with multi-layer fusion requires taking the over-fitting issue into consideration, which we discuss in the ablation study Section \ref{sec:ablation}.

\subsubsection{Representation of rotation and distance function}
To achieve continuity in the representation of rotation in deep learning \cite{continuity} is a common goal. However, to the best of our knowledge, there is no definitive conclusion in the odometry estimation task about which representation is the best, as shown in Table \ref{relatedworks}. Unlike other 6D pose-related tasks, in odometry estimation, the network predicts the ego-motion between two consecutive input frames, which means the transformation is relatively small. The same situation applies to the selection of the distance function. Similar to DeepLO \cite{deeplo}, which evaluates the rotation error of each frame in testing, we consider these two variables and conduct comprehensive experiments with our proposed loss function to select the combination with the best experimental performance in Section \ref{sec:experiment}.

\section{Methodology}
In this section, we illustrate the proposed framework TransFusionOdom in details, the network architecture is shown in Fig. \ref{architecture}. This framework includes \textit{Multi-modal} input (LiDAR and IMU), \textit{Multi-scale} modality tokens, \textit{Multi-attention} fusion (SMAF and Transformer-encoder), \textit{Multi-layer} fusion and \textit{Multi-task} regressor. 

\subsection{LiDAR and IMU data pre-processing}

The input modalities are LiDAR 3D point cloud and IMU signal. Since we use ResNet34 and ResNet18 \cite{resnet} CNN-based backbones to extract features, it is necessary to project 3D point cloud onto 2D plane $\mathbb{R}^{3} \Rightarrow \mathbb{R}^{2}$, and convert IMU signal $\mathbb{R}^{6}$ to image as well. At every timestamp $t$, the consecutive LiDAR point cloud $P_{t}$ and $P_{t+1}$, along with all IMU measurement taken between them $I_{IMU}^{t}$ are fed into the TransFusionOdom as input.

\subsubsection{Vertex and normal map of LiDAR}

Each LiDAR point cloud $p_t = (p_{x}^{t},p_{y}^{t},p_{z}^{t})$ is projected to 2D $(u,v)$ through a spherical projection method proposed in the studies of RangeNet \cite{rangenet++}, DeepLO \cite{deeplo}, DeepLIO \cite{deeplio}, and UnDeepLIO \cite{undeeplio}. The projection can be calculated as follows to obtain the vertex maps $\textit{\textbf{V}}$:
\begin{equation}
\left ( \begin{matrix}
 u\\
v
\end{matrix} \right ) =\begin{pmatrix} ({f_{u} - arctan(\frac{p_{y}}{p_{x}} ) })/{\eta_{u}} 
 \\
({f_{v} - arctan(\frac{p_{z}}{d} ) })/{\eta_{v}}
\end{pmatrix}
\label{eq1}
\end{equation}
where $d =\sqrt{p_{x}^2 + p_{y}^2 + p_{z}^2} $ denotes depth, $f_{u}$ and $f_{v}$ represent the horizontal and vertical angle in maximum. $\eta_{u}$ and $\eta_{v}$ are the resolutions of pixel representation in horizontal and vertical respectively. This mapping somehow circumvent the problem that point clouds are cluttered and disordered \cite{undeeplio}.

The normal vector is useful for point cloud registration \cite{normalpoint}. Like related works \cite{deeplio} \cite{deeplo}\cite{undeeplio}\cite{lonet}, we implement the normal maps $\textit{\textbf{N}}$, which has the correspondent relation with vertex map $\textit{\textbf{V}}$ in image coordinates ($v_{p} \Rightarrow n_{p}$). Each normal vector of point $p$ is given as bellow:
\begin{equation}
n_{p} = \sum_{i\in [0,3]}w_{p_{i},p}(v_{p_{i}} -v_{p})\times w_{p_{i+1},p}(v_{p_{i+1}} -v_{p})
\label{eq2}
\end{equation}
where $w_{m,n} = e^{-0.5\left | d(v_{a}) -d(v_{b}) \right | }$ is a pre-defined weight, $p_{i}$ denotes the four-point neighbour points in up/right/down/left directions.

\subsubsection{IMU signal image}

The mainstream approach to process IMU data is to use the recurrent neural network LSTM, which is good at modeling sequential data, as proposed in Son \emph{et al.} \cite{sonsynthetic}, UnDeepLIO \cite{undeeplio}, DeepLIO \cite{deeplio}, Li \emph{et al.} \cite{li2019deep}, Chen \emph{et al.} \cite{selectivesensor}, etc. However, LSTM has limitations in parallel computation \cite{emavio} compared with CNN-based networks. Additionally, \cite{cnnvslstm} explained that CNN-based methods are more robust than LSTM and require less time to learn the model.

Inspired by some human action recognition tasks \cite{human} \cite{human2}, we convert the IMU signal to images. This pre-processing is also necessary since we implement a multi-layer fusion strategy, unlike CNN-based backbones, LSTM does not have intermediate outputs. Before feeding the data into the TransFusionOdom framework, we conduct denoising of the IMU signal because M. Brossard \emph{et al.} \cite{denoising} stated that denoising IMU bias and noise could improve the accuracy of state estimation. Similar to VIIONet \cite{viionet}, we apply the Savitzky-Golay filter \cite{sgfilter} to filter the IMU high frequency noise. 

\begin{figure}[!t]
\centerline{\includegraphics[width=\columnwidth]{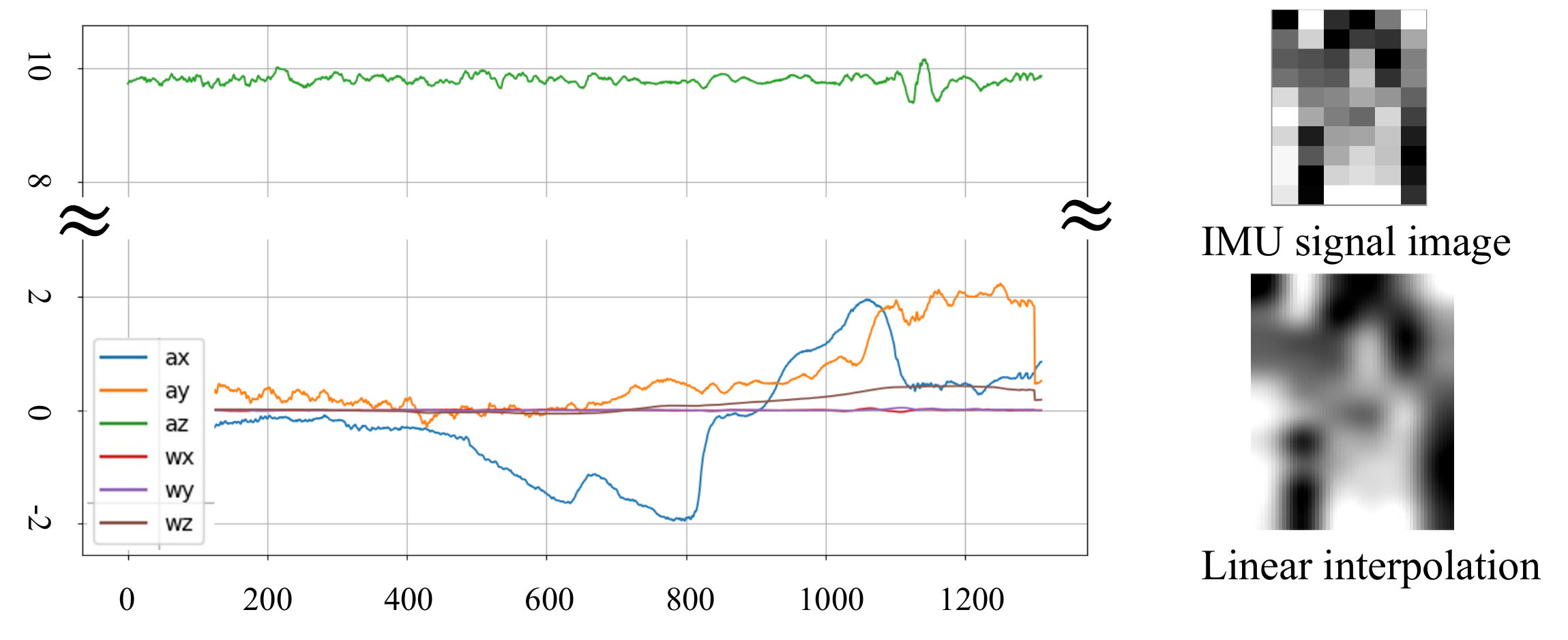}}
\caption{Raw IMU signal plotted in linear acceleration and angular velocity, IMU signal image resized with linear interpolation.}
\label{imuimage}
\end{figure}

We extract the linear acceleration and angular velocity in $x$, $y$ and $z$ axis, as shown in Fig. \ref{imuimage}. Because the frequency of IMU is higher than LiDAR, we assume there are $\gamma $ IMU measurements between $I_{LiDAR}^{t}$ and $I_{LiDAR}^{t+1}$. The raw IMU signal image $I_{IMU}^{t,t+1}$ between timestamp $[t, t+1]$ is generated as follows: 
\begin{equation}
I_{IMU}^{t,t+1}=\left [ \begin{matrix}
 IMU_{0}\\
 IMU_{1}\\
 IMU_{2}\\
 ...\\
IMU_{\gamma }
\end{matrix} \right ]  = \begin{bmatrix}
 a_{x}^{0}, a_{y}^{0}, a_{z}^{0} , w_{x}^{0} ,w_{y}^{0}  ,w_{z}^{0} \\
  a_{x}^{1}, a_{y}^{1} ,a_{z}^{1}  ,w_{x}^{1}  ,w_{y}^{1}  ,w_{z}^{1} \\
  a_{x}^{2},  a_{y}^{2} , a_{z}^{2}  ,  w_{x}^{2} ,  w_{y}^{2} , w_{z}^{2} \\
 ... \\
   a_{x}^{\gamma }, a_{y}^{\gamma } ,a_{z}^{\gamma },w_{x}^{\gamma }  ,w_{y}^{\gamma }  ,w_{z}^{\gamma }
\end{bmatrix}
\label{eq3}
\end{equation}

To avoid excessive information compression, particularly in the width dimension, when conducting multi-layer fusion using intermediate outputs of the ResNet model with the LiDAR modality, we normalize and resize the original IMU signal image, which is $\in \mathbb{R}^{p \times 6 \times 3}$, to $(H_{IMU},W_{IMU})$. We conducted the experiments to compare the performance using learnable layer to enlarge the image and a fixed linear interpolation. The results show that the latter solution is better, possibly because the learnable layer has more uncertainty in extracting representative features at the following steps.

\subsection{Soft-mask attentional fusion for homogeneous data}

After the initial ResNet layer, two backbones share the same weights to extract common features from the vertex and normal maps ($V_p$ and $N_p$). This weight sharing can help compress the size of the model. Because $V_p$ and $N_p$ have the same shape and a certain corresponding relationship exists between each element during pre-processing. The fusion of $V_p$ and $N_p$ is defined as homogeneous multi-modal fusion \cite{channelexchange}.

We conduct the SMAF during each layer of ResNet, which is a MLP-based learnable mask. Similar attention mechanisms are introduced in \cite{atvio}, \cite{selectfusion}, \cite{pointloc} and \cite{unsuperviseddeeplo}. Here $v_{p}^{l_{1}}$ and $n_{p}^{l_{1}}$ are the output of the first layer in ResNet34, MLP used to convert features generated by concatenation $[v ;n]$ to mask vector $M_{vertex}^{l_{1}}$ and $M_{normal}^{l_{1}}$, $Sigmoid$ is used to reweight the mask to the range of $[0,1]$. The whole processing is automatically parameterized by network as follows:
\begin{equation}
\begin{matrix}
M_{vertex}^{l_{1}}=Sigmoid(MLP_{v}[v_{p}^{l_{1}};n_{p}^{l_{1}}])
 \\
M_{normal}^{l_{1}}=Sigmoid(MLP_{n}[v_{p}^{l_{1}};n_{p}^{l_{1}}])
\end{matrix}
\label{eq4}
\end{equation}

Utilizing the mask vectors, the input homogeneous modalities vertex and normal maps are reweighed by element-wise multiplication $\otimes$. This SMAF $\tau _{soft}(v,n)$ operation conducts at layer $t$ of ResNet, modelled as:
\begin{equation}
\tau _{soft}(v_{p}^{l_{t}},n_{p}^{l_{t}})=v_{p}^{l_{t}}\otimes M_{vertex}^{l_{t}}+ n_{p}^{l_{t}}\otimes M_{normal}^{l_{t}}
\label{eq5}
\end{equation}

\subsection{Transformer-based fusion for heterogeneous data}
The output of SMAF $\tau _{soft}(v_{p}^{l_{t}},n_{p}^{l_{t}})$ needs to be fused with the IMU modality $i_{p}^{l_{t}}$ after each layer of ResNet, which is a heterogeneous multi-modal fusion task. In this article, we utilize the Transformer to perform heterogeneous data fusion. Unlike LSTM, Transformer can perform parallel computation, but the computational complexity of self-attention is quadratic in the number of tokens \cite{vit}. Especially when using a multi-layer strategy, it is important to reduce the size of tokens to maintain the efficiency and size of the entire model.

\begin{figure}[!t]
\centerline{\includegraphics[width=\columnwidth]{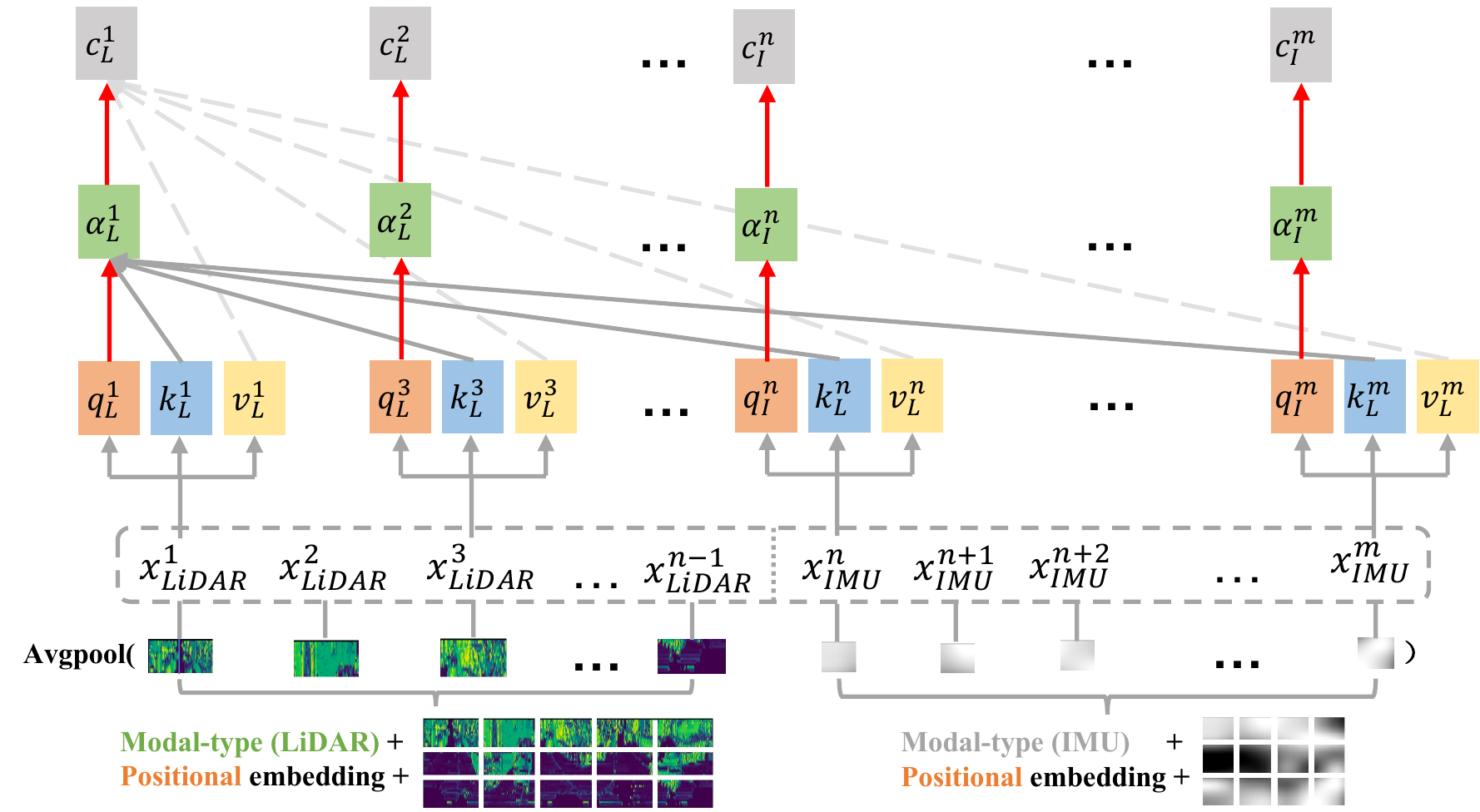}}
\caption{Transformer-based fusion between LiDAR and IMU tokens.}
\label{encoder}
\end{figure}

Different from ViT\cite{vit} and ViLT \cite{vilt}, which use image patches as token, we apply the average pooling operation to down-sample the image patch and obtain a set $\in\mathbb{R}^{H \times W \times C}$, which includes $[x_{L}^{1},x_{L}^{2},x_{Li}^{3},...,x_{L}^{n-1}]$ and $[x_{I}^{n},x_{I}^{n+1},x_{I}^{n+2},...,x_{I}^{m}]$, In this set, each element is treated as a token. We avoid the disorder and clutter of point clouds by positional encoding $x_{l}^{pos}/ x_{i}^{pos}$ $\in \mathbb{R}^{H \times W \times C} $, in addition to project raw point cloud to image. We also apply the modal-type embedding $ l^{type}/i^{type} \in \mathbb{R}^{W}$ to give prior knowledge about each token belongs to which modality. The modal-type embedding is generated by a learnable linear layer. The ability of differentiation between different sources could improve the performance of Transformer, which has been validated in AFT-VO \cite{aftvo} and ViLT \cite{vilt}. The set sequence, positional embedding and modal-type embedding of each modality integrate together by element-wise summation as follows:
\begin{equation}
\bar{x}_{LiDAR}=[x_{L}^{1},x_{L}^{2},x_{Li}^{3},...,x_{L}^{n-1}] + x_{l}^{pos}
\label{eq9}
\end{equation}
\begin{equation}
\bar{x}_{IMU}=[x_{I}^{n},x_{I}^{n+1},x_{I}^{n+2},...,x_{I}^{m}] + x_{i}^{pos} 
\label{eq10}
\end{equation}
\begin{equation}
 \mathbf{G}^{in}=[\bar{x}_{LiDAR}+l^{type};\bar{x}_{IMU}+i^{type}]
\label{eq11}
\end{equation}
The input to Transformer-encoder is $\mathbf{G}^{in}$$\in \mathbb{R}^{M \times D_{f}}$, each token is a feature vector with dimension of $D_{f}$. The query $\mathbf{Q}$, key $\mathbf{K}$ and value $\mathbf{V}$ are generated by linear transformation with $\mathbf{M} ^{q}$ $\in \mathbb{R}^{D_{f} \times D_{q}} $, $\mathbf{M} ^{k}$ $\in \mathbb{R}^{D_{f} \times D_{k}} $ and $\mathbf{M} ^{v}$ $\in \mathbb{R}^{D_{f} \times D_{v}} $, respectively:
\begin{equation}
\mathbf{Q}  = \mathbf{G} ^{in} \mathbf{M} ^{q}, \mathbf{K} =\mathbf{G} ^{in}\mathbf{M} ^{k},\mathbf{V} =\mathbf{G} ^{in}\mathbf{M} ^{v}
\label{eq6}
\end{equation}
Then the attention mechanism shown in Fig. \ref{encoder} inside of transformer encoder is calculated by the following formulas:
\begin{equation}
\bm{\alpha}_{L,I}=\frac{\mathbf{QK} ^{T}}{\sqrt{D_{k}} } 
\label{eq12}
\end{equation}
\begin{equation}
\mathbf{C}_{L,I}=softmax(\bm{\alpha}_{L,I})\mathbf{V} 
\label{eq7}
\end{equation}
\begin{equation}
\mathbf{G} ^{out} = MLP(\mathbf{C} )+\mathbf{G} ^{in}
\label{eq8}
\end{equation}
where $\mathbf{G} ^{out}$ is the same shape with $\mathbf{G} ^{in}$. There are several layers which are applied in original Transformer-encoder, the multi-heads attention generates parallel $\mathbf{Q,K,V}$ and involves concatenating the attention value of $\mathbf{C_{L,I}}$.

The output of the fusion, $\mathbf{G^{out}}$, is up-sampled to recover to the original resolution through bilinear interpolation, which is the dimension of each layer's output from ResNet. After up-sampling, element-wise summation is used to integrate the output with existing feature maps as residual learning \cite{resnet} to prevent gradient degradation.

Different from TransFuser \cite{transfuser}, which is scaled the tokens at a fixed shape $H_{0}=H_{1}=H_{2}=H_{n}, W_{0}=W_{1}=W_{2}=W_{n} $ for each layer, we gradually scaled the dimensions of tokens for each layer, correspondent with the shape of each layer in ResNet, instead of a fixed resolution of tokens for multi-layer fusion, so that we have a multi-scale Transformer tokens fusion strategy. The advantages of multi-scale Transformer have been highlighted in many works \cite{multiscale1, multiscale2, multiscale3, general}, which leverage the coarse-to-fine concept of the classic image pyramid architecture \cite{coarse1}. 

\subsection{Interpretable multi-modal fusion inside of Transformer}
\begin{figure}[!t]
\centering
\subfigure[The generation of attention matrix and the definition of self/cross attention component.]{
\includegraphics[width=\columnwidth]{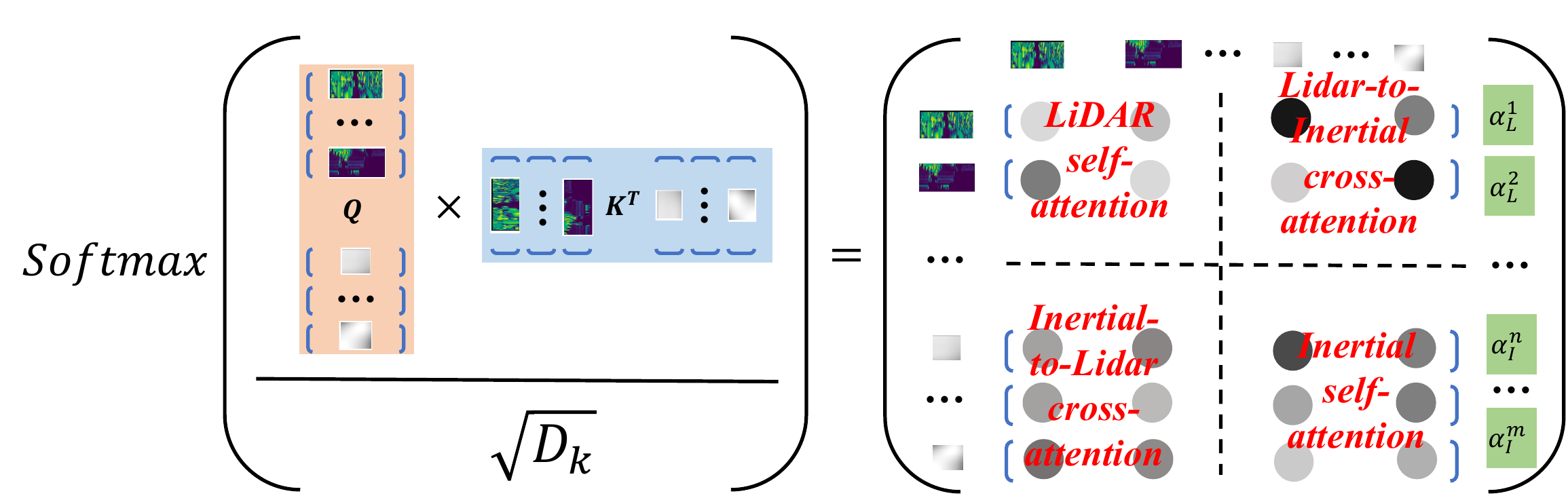}
\label{matrixdef}
}
\quad
\subfigure[The generation of attention map and the relationship between query tokens and attended tokens with highest score.]{
\includegraphics[width=\columnwidth]{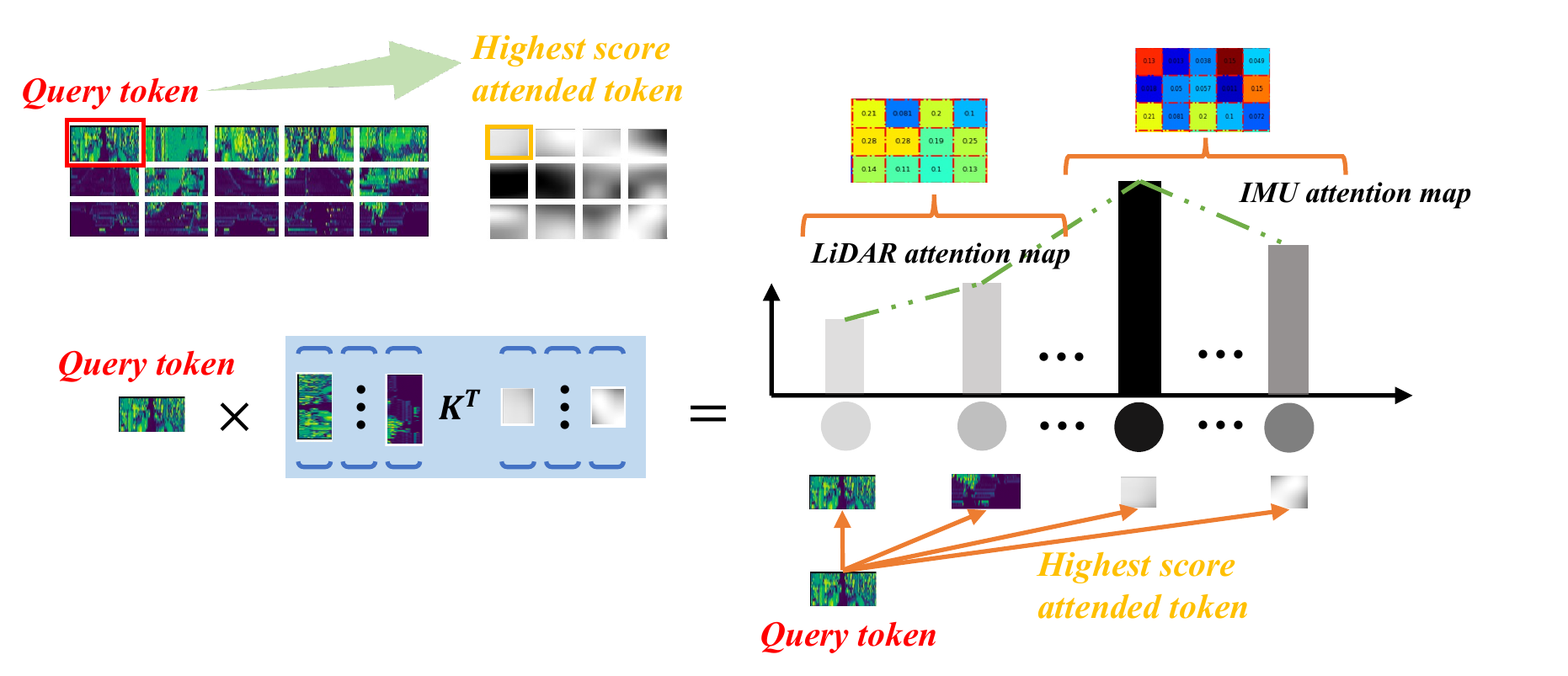}
\label{attendedtoken}
}
\caption{The interpretation of interactions between LiDAR and IMU inside the Transformer.}
\label{attention}
\end{figure}

In this subsection, we demonstrate the interpretation of interactions between two modalities $\bar{x}_{LiDAR}$ and $\bar{x}_{IMU}$ inside the Transformer-encoder. Through the generation of the attention matrix $\bm{\alpha}_{L,I}$ $\in \mathbb{R}^{m \times m} = [\alpha_{L}^{1}; \alpha_{L}^{2};...;\alpha_{I}^{m}]$ , it can be split into four components which represent different self and cross-attentions. If the query and key belong to the same modality, such as $(x_{L}^{n}\dagger x_{L}^{n'})$ and $(x_{I}^{n}\dagger x_{I}^{n'})$, where ${\dagger}$ denotes the process of using the query to search the key, we define it as self-attention. Otherwise, $(x_{L}^{n}\dagger x_{I}^{n'})$ is defined as cross-attention. Based on this definition, in Fig. \ref{matrixdef}, the top-left of $\bm{\alpha}_{L,I}$ represents LiDAR self-attention, the top-right represents LiDAR-to-IMU cross-attention, the bottom-left represents IMU-to-LiDAR cross-attention, and the bottom-right represents IMU self-attention.

Each column in $\bm{\alpha}_{L,I}$ is the attention weights of one query token in two modalities. We can reshape the weights to their original resolution of corresponding modality. The relationship between query token and attended token with highest score could be observed as shown in Fig. \ref{attendedtoken}. The detailed visualization results are discussed in Section IV.B. 

\subsection{Multi-task regression}
This section is the extension of our previous work CertainOdom \cite{certianodom}. In the supervised 6D pose regression field, the objective is to predict the 6D pose vector $\mathbf{p}$ including translation and rotation $x$ and $q$.
\begin{equation}
\mathbf{p} =[x,q]
\label{eq14}
\end{equation}

The baseline method proposed in PoseNet \cite{posenet} and employed in DeepVO \cite{deepvo}, ATVIO \cite{atvio}, VINet \cite{vinet} and SelectFusion \cite{selectfusion} is as follows,
\begin{equation}
loss = \left \| x- \hat{x}   \right \| _{2}  + \beta \left \| \frac{q}{\left \| q \right \|  } -\hat{q} \right \| _{2}
\label{eq15}
\end{equation}
where L2 loss is used as the distance function to calculate the error between prediction and ground truth $[\hat{x},\hat{q}]$, the manually tuned hyperparameter $\beta$ is used to weigh the error between translation and rotation.

To overcome the problems comes from hyperparameters, DeepLO\cite{deeplo}, Lo-Net \cite{lonet} and MS-Transformer \cite{MS-transformer} introduce the following loss function:
\begin{equation}
loss = L_{x}exp(-s_{x}) + s_{x} +L_{q}exp(-s_{q}) + s_{q}
\label{eq16}
\end{equation}
where $L_{x,q}$ denotes the distance function, $s_{x}$ and $s_{q}$ are learned parameters to balance the error between translation and rotation.


The main contribution of our previous work CertainOdom \cite{certianodom} is to leverage the uncertainty which regressed from multi-decoders as multi-task learning to weigh the error from translation and rotation automatically. The supervised uncertainty estimation proposed by Kendall \emph{et al.} \cite{whatuncertainty} is as follows,
\begin{equation}
loss = \frac{1}{N} \sum_{i=1}^{N} \frac{1}{2\sigma (x_{i} )^{2}} \left \| y_{i}-f(x_{i}) \right \|^{2} + \frac{1}{2}log\sigma (x_{i})^{2}  
\label{eq17}
\end{equation}
where $\sigma$ is the predicted aleatoric uncertainty for the input $x$; $y_{i}$ and $f(x_{i})$ denote the ground truth and prediction with regard to input $x_{i}$ respectively.

Besides, similar with CertainOdom\cite{certianodom}, DeepLIO \cite{deeplio}, VINet \cite{vinet}, EMA-VIO \cite{emavio} and what Zou \emph{et al.} proposed \cite{f2ff2g}, the frame-to-frame (f2f) constraint represents the relative transformation between each frame in one sliding window and frame-to-global (f2g) is an absolute pose denotes each the transformation from each frame to initial frame. 

Regarding the selection of rotation representation, VINet \cite{vinet} and DeepLIO \cite{deeplio} incorporate the Lie algebra $se(3)$ and Lie group $SE(3)$ into the f2f and f2g constraints, respectively, which yield better performance than using a unique representation of rotation. The definitions are as follows:

\begin{equation}
\mathbf{se(3)} = \left \{ \xi \in \mathbb{R} ^6 = \left [ \binom{\rho }{\phi }  \right ], \rho \in \mathbb{R} ^3,\phi \in so(3)   \right \} 
\label{eq9_}
\end{equation}

\begin{equation}
\begin{aligned}
\mathbf{SE(3)} =  \mathbf{T} \in \mathbb{R}^{4\times 4} &=\left \{ \begin{pmatrix}
 R &t \\
 0^{T} &1
\end{pmatrix} \mid  R  \in SO(3), t\in \mathbb{R}^{3\times 1}  \right \} , \\ \hat{\xi} \in \mathbb{R} ^{4\times 4} &=\begin{bmatrix}
 \hat{\phi } & \rho \\
  0^{T}&0
\end{bmatrix} 
\label{eq18}
\end{aligned}
\end{equation}

However, it is difficult to directly regress the rotation matrix $ R  \in SO(3), R\in \mathbb{R}^{3\times 3}$, since it needs to enforce their special orthogonal properties \cite{gabasape}. There are 3 options for the conversion of rotation matrix, as we listed in Table \ref{relatedworks}, Euler angle, quaternion and axis angle. Different from our previous work, we conduct the exhaustive experiments to obtain the best combination between different representation of rotation and distance functions in Section IV.B. 


The predictions for f2f and f2g are represented as $y_{i}^{f}=(\rho_{i}^{f},\phi_{i}^{f})$ and $y_{i}^{g}=(t_{i}^{g},\varphi_{i}^{g})$, where $\varphi_{i}^{g}$ denotes Euler angles. The corresponding ground truths are represented as $\hat{y} {i}^{f} $ and $\hat{y} {i}^{g} $. Additionally, we formulate the proposed loss $L(\theta,\sigma_{x},\sigma_{y},\sigma_{z},\sigma_{r_{x}},\sigma_{r_{y}},\sigma_{r_{z}})$ as a function of our network, where $\theta$ represents the weights of the network and $\sigma(x,y,z,r_{x},r_{y},r_{z})$ denotes the uncertainties in translation and rotation. Combining the supervised baseline loss and uncertainty estimation regression loss, the loss can be calculated as follows:
\begin{equation}
\begin{aligned}
L_{x}(\theta ,\sigma_{x}) =  
\textstyle \sum_{i}^{}\frac{1}{2} exp(-s_{x}^{i} )([ \left \| \rho _{ix}^{f} -\hat{ \rho}  _{ix}^{f}   \right \|^{2},\\ \left \| t _{ix}^{g} -\hat{ t}  _{ix}^{g}   \right \|^{2} ]) +\frac{1}{2}s_{x}^{i}  
\label{eq19}
\end{aligned}
\end{equation}

\begin{equation}
\begin{aligned}
L_{r}(\theta ,\sigma_{r_{x}}) =  \textstyle \sum_{i} \frac{1}{2} exp(-s_{r_{x}}^{i} )([\left \| \phi _{ir_{x}}^{f} -\hat{\phi}  _{ir_{x}}^{g}   \right \|^{2}, \\ \left \| \varphi_{ir_{x}}^{f}  -\hat{ \varphi}  _{ir_{x}}^{g}   \right \|^{2} ] ) +\frac{1}{2}s_{r_{x}}^{i}  
\label{eq20}
\end{aligned}
\end{equation}
Where $\rho_{i}^{f}\in \mathbb{R}^{3}=(\rho_{ix}^{f},\rho_{iy}^{f},\rho_{iz}^{f})$ represents the translation in the x, y, and z axis, and $\varphi_{i}^{f}\in \mathbb{R}^{3}=(\varphi_{ir_{x}}^{f},\varphi_{ir_{y}}^{f},\varphi_{ir_{z}}^{f})$ represents the rotation in roll, pitch, and yaw. Following the approach of \cite{whatuncertainty}, we use log variance $s_{i}=log\sigma_{i}^{2}$ for uncertainty implementation. By using learnable uncertainties, the error between translation and orientation can be automatically weighted, rather than using pre-defined weighting hyperparameters, which is similar to how Kendall \emph{et al.} \cite{multitask} use uncertainty to weigh semantic, instance segmentation, and depth estimation tasks in a multi-task learning framework. Finally, the joint loss can be computed as following:
\begin{equation}
\begin{aligned}
L(\theta,\sigma _{i}^{x,y,z,r_{x},r_{y},r_{z}}) =   {\textstyle \sum_{i}^{}}  (L_{x}(\theta ,\sigma_{i}^{x})+...+L_{r_{z}}(\theta ,\sigma_{i}^{r_{z}}))
\label{eq21}
\end{aligned}
\end{equation}

\section{Experiments}
\label{sec:experiment}
 
The proposed TransFusionOdom is implemented with Pytorch and 250 epochs are trained with NVIDIA Quadro GV100 (32GB VRAM). The inference fps is around 20, which meets some real-time applications. The window size of f2f and f2g constraints is 8 and batch size is set to 16. Multi-layer fusion is conducted 4 times after the repeated conventional layers of ResNet34 (3,4,6,3) and ResNet18 (2,2,2,2). The number of multi-head is set to 4. Most of empirical hyperparameters are referenced from other related works, such as MLF-VO \cite{mlfvo} and Transfuser \cite{transfuser}.

In this section, we first conduct experiments to demonstrate how to select the best combination of different representations of rotation and distance functions for the proposed loss function. Then, we visualize the SMAF and Transformer-based attention weights. Through visualization, we can observe the interactions between multiple modalities. The visualization of different attentional mechanisms inside the fusion process improves the interpretation of the proposed framework and convinces us the competitive performance of the proposed method.

For the dataset, KITTI \cite{kitti} is an autonomous driving dataset commonly used as a benchmark for evaluating odometry tasks. We use sequence 00-08 for training and 09-10 for validation. As mentioned in \cite{deeplio}, since IMU and LiDAR are not synchronized, the number of IMU measurements between two LiDAR frames varies, with 10 to 13 IMU measurements between two consecutive LiDAR frames. We normalized the IMU measurements and set $\gamma = 10$, as introduced in Section III.A.2. The overfitting problem in trajectory estimation and uncertainty evaluation are tested on KITTI dataset. All statistical metrics are calculated by the public available KITTI evaluation tool \footnote{\url{https://github.com/LeoQLi/KITTI_odometry_evaluation_tool}}. 

We also publish a synthetic multi-modal dataset for odometry estimation based on the Gazebo simulation environment. This dataset can be used to conveniently validate the generalization ability of the proposed fusion strategy.
\begin{table}[!t]
\caption{All combinations of representation of rotation with L1/2 distance }
\label{combinations}
\renewcommand\arraystretch{1.5}
\begin{center}
\scalebox{0.8}{
\begin{tabular}{c|c}
\hline
&                                                                 \\
\multirow{-2}{*}{} & \multirow{-2}{*}{{ Rotation combinations with L1 and L2 distance}} \\ \hline
1                  & { Both Euler angle}                                                \\ \cline{1-1}
2                  & { Both Axis-angle}                                                 \\ \cline{1-1}
3                  & { Both Quaternion}                                                 \\ \cline{1-1}
4                  & { Both se3}                                                        \\ \cline{1-1}
5/6                & { Euler (f2f) + Axis-angle (f2g) / Opposite}                       \\ \cline{1-1}
7/8                & { Euler (f2f) + quaternion (f2g) / Opposite}                       \\ \cline{1-1}
9/10               & { Euler (f2f) + se3 (f2g) / Opposite}                              \\ \cline{1-1}
11/12              & { Axis-angle (f2f) + quaternion (f2g) / Opposite}                  \\ \cline{1-1}
13/14              & { Axis-angle (f2f) + se3 (f2g) / Opposite}                         \\ \cline{1-1}
15/16              & { Quaternion (f2f) + se3 (f2g) / Opposite}                         \\ \hline
\end{tabular}
}
\end{center}
\end{table}

\subsection{Representations of rotation and distance functions}
Since we consider that the selection of the representation of rotation and distance function is sensitive to loss functions and there is no definitive conclusion about which one is the best for learning-based odometry estimation, as mentioned in Table \ref{relatedworks}, we conduct experiments on all combinations with f2f and f2g constraints, which are listed in Table \ref{combinations}.

\begin{figure}[t]
\centerline{\includegraphics[width=6.5cm]{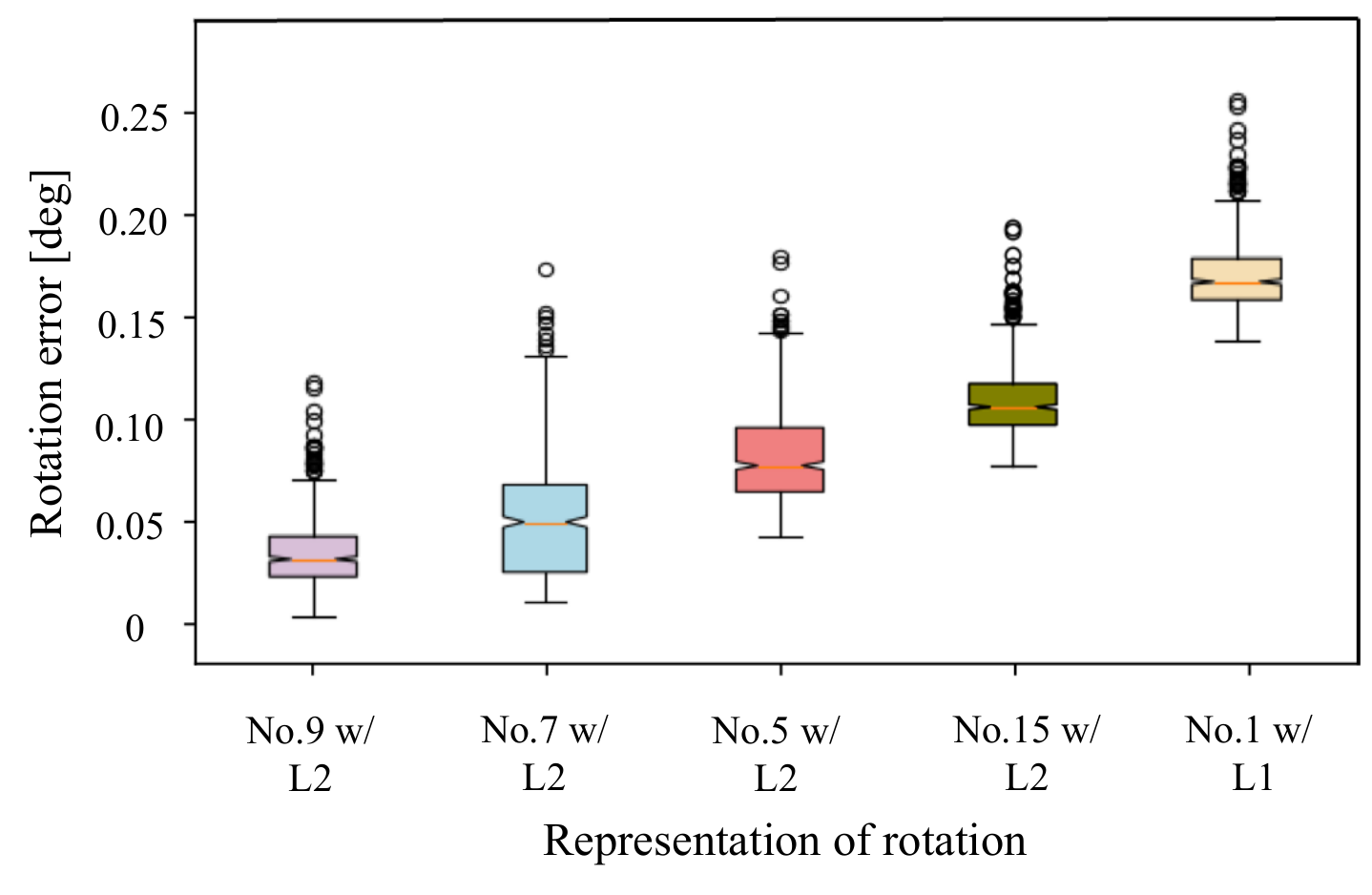}}
\caption{Box plot about the top-5 combinations of representation of rotation and distance function using rotation error.}
\label{boxplot}
\end{figure}

The approach for selecting the best one is to calculate the rotation error under the same training situation, different from directly plot the rotation error in DeepLO \cite{deeplo}, we use box plot for observation. Through the box plot in Fig. \ref{boxplot}, we can observe the mean rotation error, stability and outliers of different combinations. The best one with our proposed loss is Euler angle with f2f constraint and se3 with f2g, combined with L2 distance function. Based on all results, we find out that Euler angle is more suitable for relative transformation and se3 is good at global transformation, L2 is better than L1 in most of situations. However, compared with the conclusion we obtain through extensive experiments, the approach used to evaluate all combinations is more useful in case we need to validate with other loss functions for different tasks.

\begin{figure}[t]
\centering
\subfigure[SMAF between vertex and normal map during straight and turning, the car starts to turn around $70^{th}$ and finish in the end.]{
\includegraphics[width=\columnwidth]{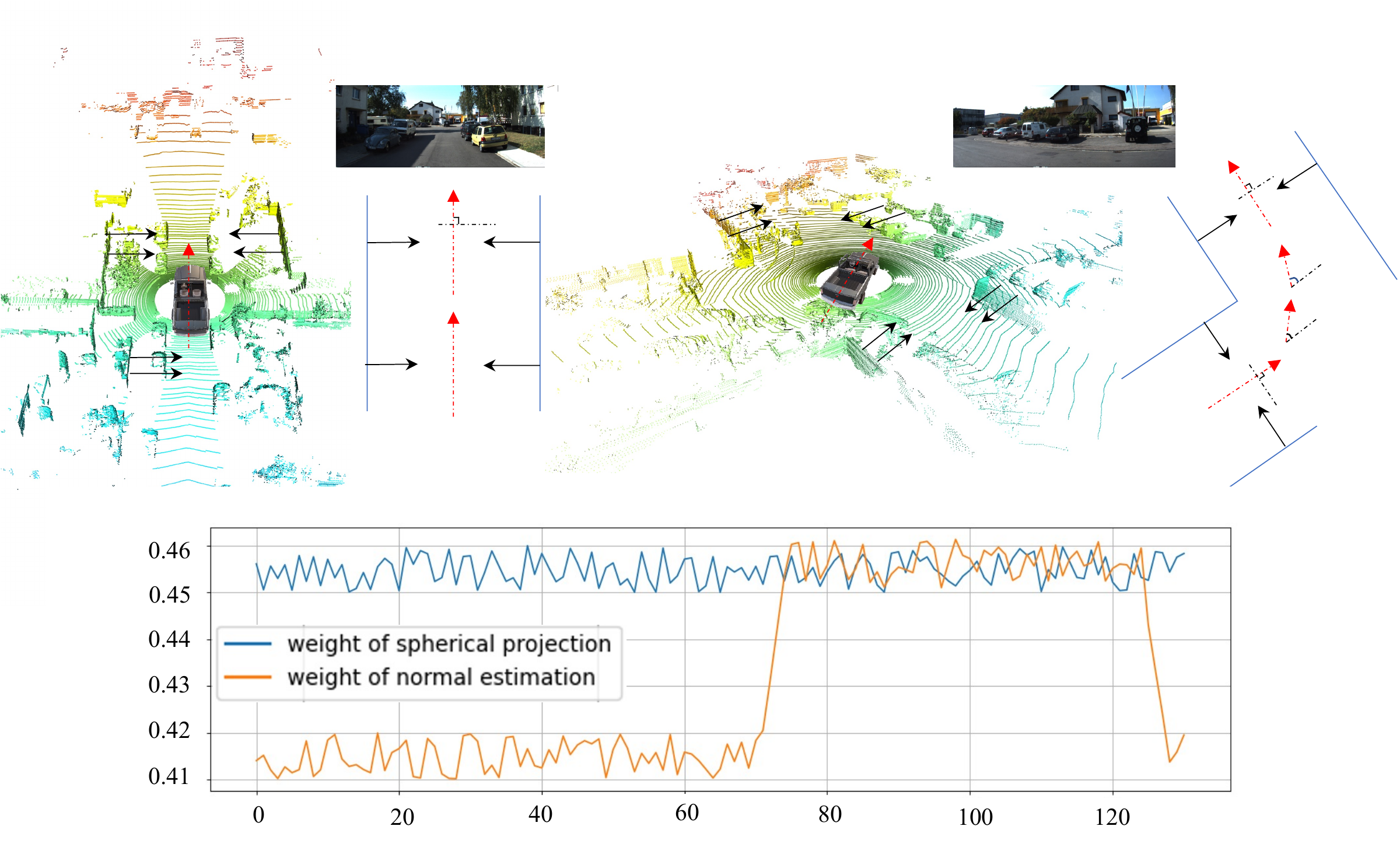}
\label{weights}

}
\quad
\subfigure[Normal estimation under different (left: forest, right: wall) road situations.]{
\includegraphics[width=7.5cm]{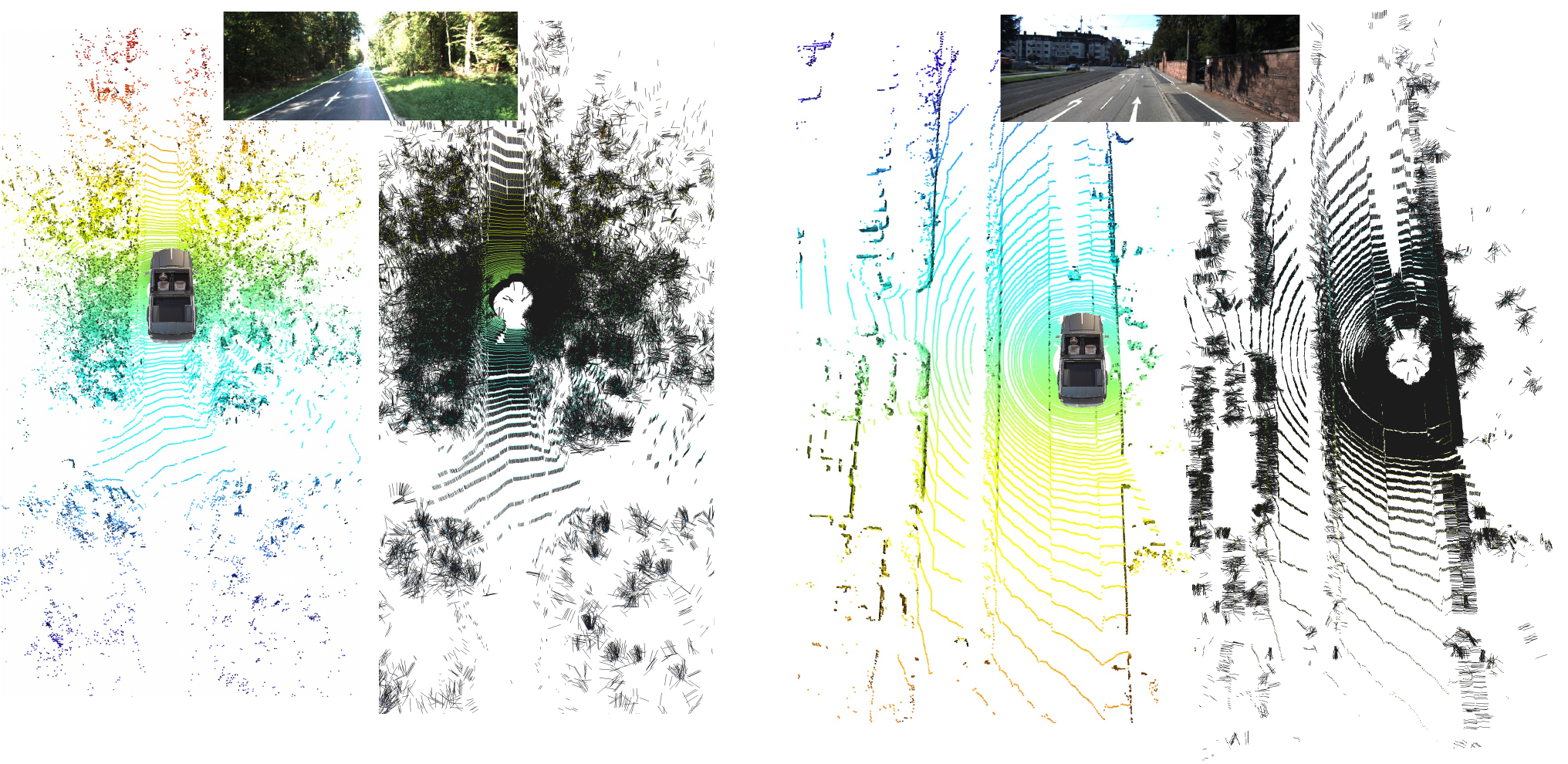}
\label{normal}
}
\caption{Visualization of attention between homogeneous modalities}
\end{figure}

\subsection{Visualization of interactions between multiple modalities}

\subsubsection{Attention inside of homogeneous fusion}
According to the introduction in Section III.B, we plot the soft mask weights $M_{vertex}^{l_{1}}$ and $M_{normal}^{l_{1}}$ in Fig. \ref{weights}. We observe that during turning, the weights of the normal map increased a lot compared to straight driving in $l_{1\sim4}$. The possible reason is that the angle between the driving direction and normal vector changes more during turning than during straight driving. Similar conclusions were mentioned in UnDeepLIO \cite{undeeplio}, where they only used vertex information to estimate the translation because the change in translation does not affect the normal information. However, we also find out that the normal map is sensitive to road situations, as shown in Fig. \ref{normal}. For example, when the surroundings are in a forest, the normal information is in a highly random condition compared to a wall along the driving direction. Generally, the network gives more attention to changing information instead of static features.

\subsubsection{Attention inside of heterogeneous fusion}
\begin{figure}[!t]
\centerline{\includegraphics[width=\columnwidth]{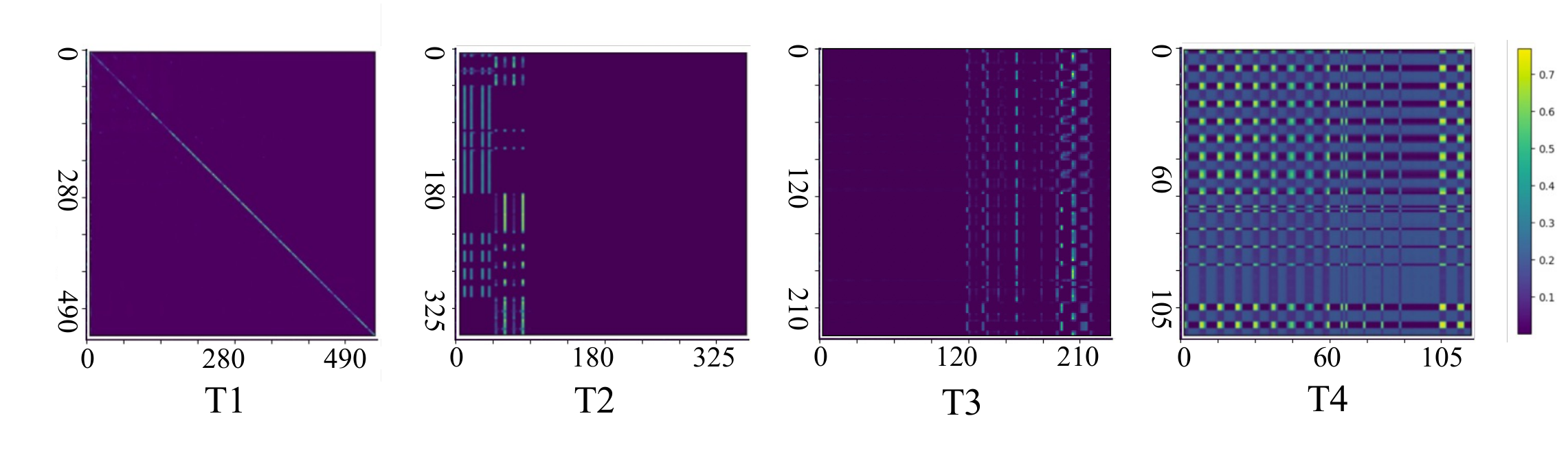}}
\caption{Attention matrix of T1 to T4 in head 4.}
\label{t14}
\end{figure}

\begin{table}[t]
\caption{Cross-attention quantitative evaluation}
\label{tab:attention statistic}
\renewcommand\arraystretch{1.5}
\begin{center}
\scalebox{0.85}{
\begin{tabular}{c|cccccccc}
\hline
\multirow{2}{*}{Head} & \multicolumn{2}{c|}{T1}                                       & \multicolumn{2}{c|}{T2}                                       & \multicolumn{2}{c|}{T3}                                       & \multicolumn{2}{c}{T4}                   \\ \cline{2-9} 
                      & \multicolumn{1}{c|}{$L_{t}$} & \multicolumn{1}{c|}{$I_{t}$} & \multicolumn{1}{c|}{$L_{t}$} & \multicolumn{1}{c|}{$I_{t}$} & \multicolumn{1}{c|}{$L_{t}$} & \multicolumn{1}{c|}{$I_{t}$} & \multicolumn{1}{c|}{$L_{t}$} & $I_{t}$ \\ \hline
1                     & 0                             & 0                             & 5.7                           & 22.3                          & 35.6                          & 15.6                          & 60.7                          & 63.5     \\ \cline{1-1}
2                     & 0                             & 0                             & 7.3                           & 25.6                          & 42.3                          & 23.6                          & 65.4                          & 67.4     \\ \cline{1-1}
3                     & 0                             & 0                             & 10.1                          & 33.2                          & 44.5                          & 22.5                          & 70.3                          & 67.9     \\ \cline{1-1}
4                     & 1.2                           & 1.5                           & 12.7                          & 30.8                          & 45.7                          & 21.3                          & 76.7                          & 80.3     \\ 
\hline
\end{tabular}
}
\end{center}
\end{table}

Based on the approach we introduced in Section III.D, we first visualize the attention matrix of head 4 from Transformer 1 to 4 as shown in Fig. \ref{t14}. By observing the position of high-value attended tokens, we can see that in T1, the self-attention domains are more prominent compared to the later Transformer fusion stages. From T2 to T4, the network gradually learns cross-attention. In the final layer T4, the highest values of attended tokens are distributed uniformly, indicating that there are interactions based on attention weights between the two modalities inside the fusion.

In addition to qualitative results, we conduct the statistical analysis of cross-attention on KITTI dataset 09 sequence as shown in Table \ref{tab:attention statistic}. We present the percentage of tokens ($L_{t}$: LiDAR tokens, $I_{t}$: inertial tokens) that have at least three of the top-5 attended tokens belonging to the other modality in each head of T1 to T4. Consistent with the visualization of the attention matrix, T1 shows rare cross-attention. Besides, in each fusion stage, almost all the later heads exhibit more cross-attention weights than their former heads. Also, the value of T4 indicates that our proposed fusion strategy is capable of aggregating information from two modalities.

\begin{figure}[!t]
    \centering
    \subfigure[Inertial-to-Inertial self attention, query token: the first patch.]{%
        \includegraphics[width=7.15cm]{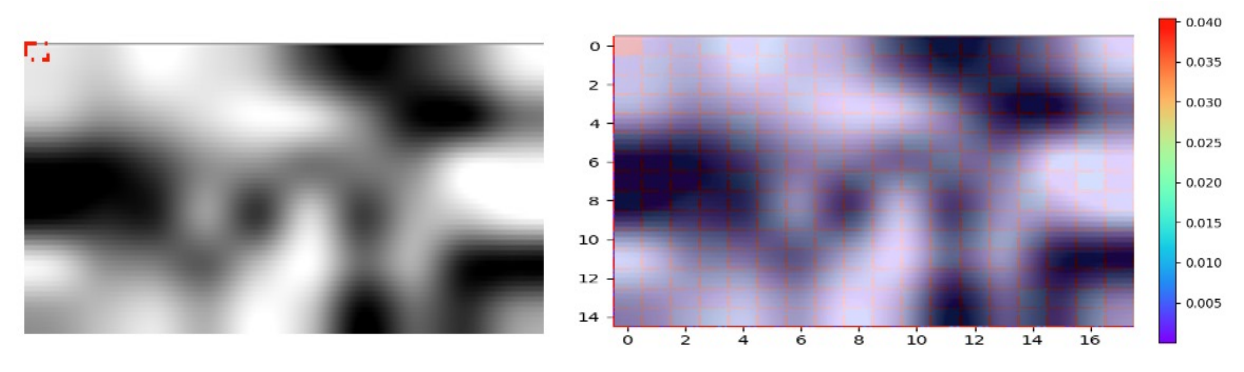}
        \label{overlap0}
    }
    
    \subfigure[Inertial-to-LiDAR cross attention, query token: high linear acceleration value.]{%
        \includegraphics[width=8.6cm]{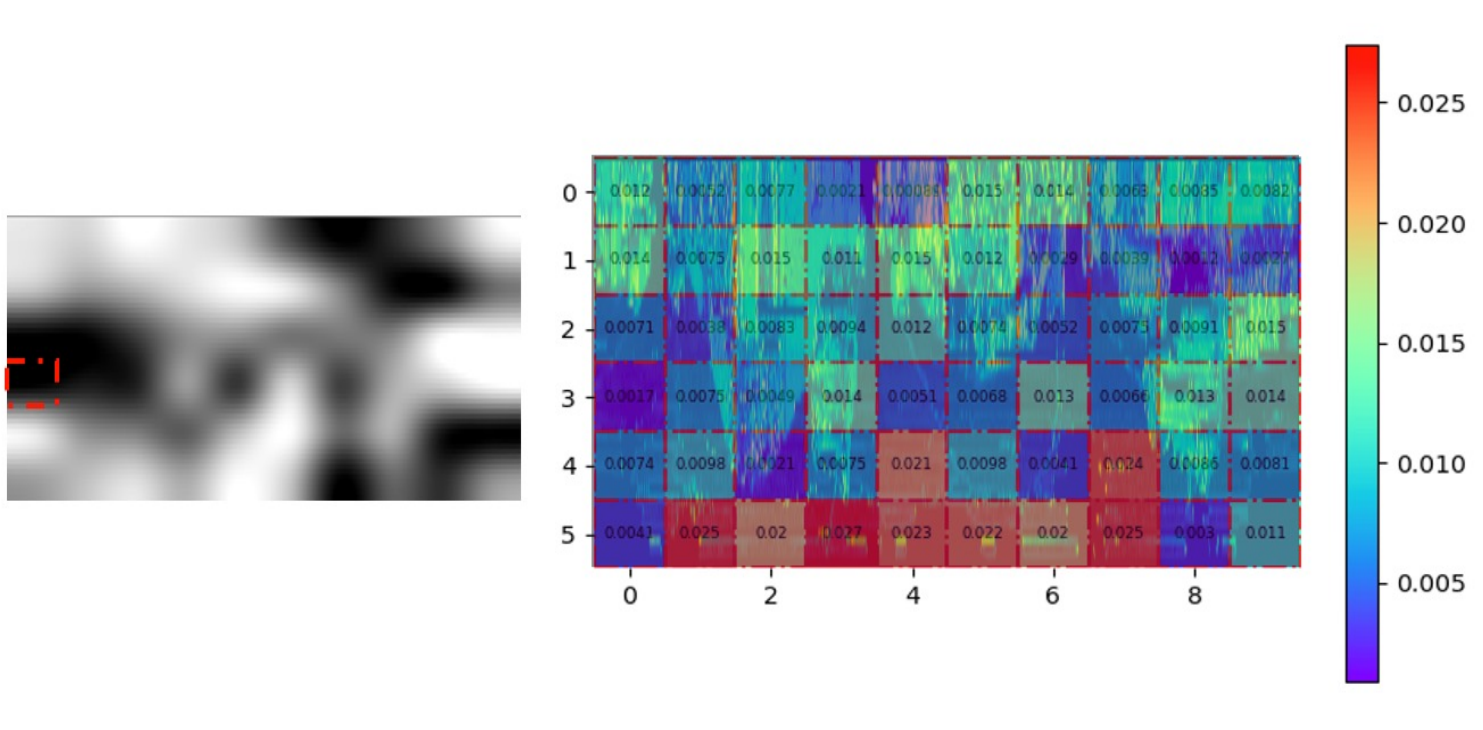}
        \label{overlap1}
    }
    
    \subfigure[Inertial-to-LiDAR cross attention, query token: low angular velocity value.]{%
        \includegraphics[width=8.6cm]{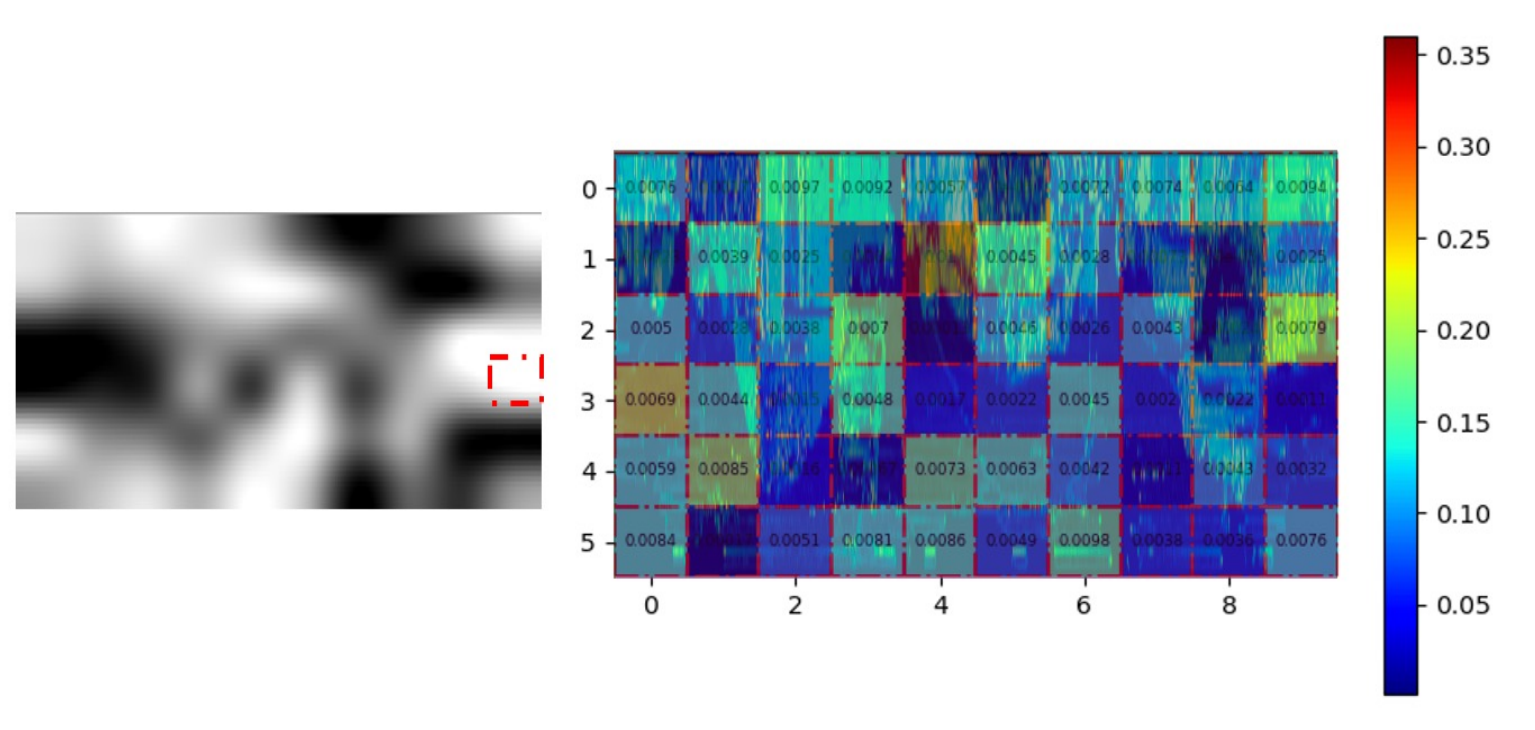}
        \label{overlap2}
    }
    
    \subfigure[Inertial-to-LiDAR cross attention, query token: high angular velocity value.]{%
        \includegraphics[width=8.6cm, height=4.4cm]{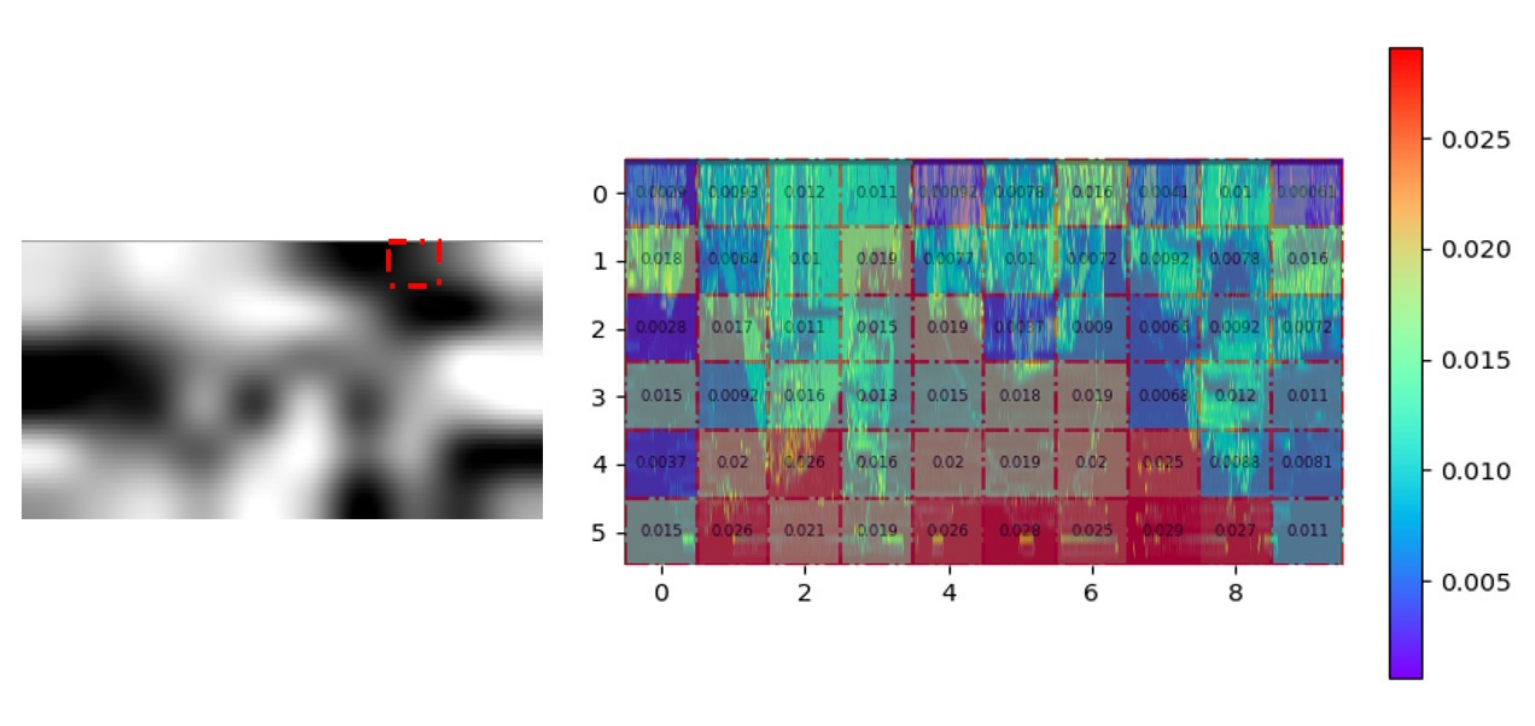}
        \label{overlap3}
    }
    
    \subfigure[LiDAR-to-Inertial cross attention, query token: corner position.]{%
        \includegraphics[width=8.4cm]{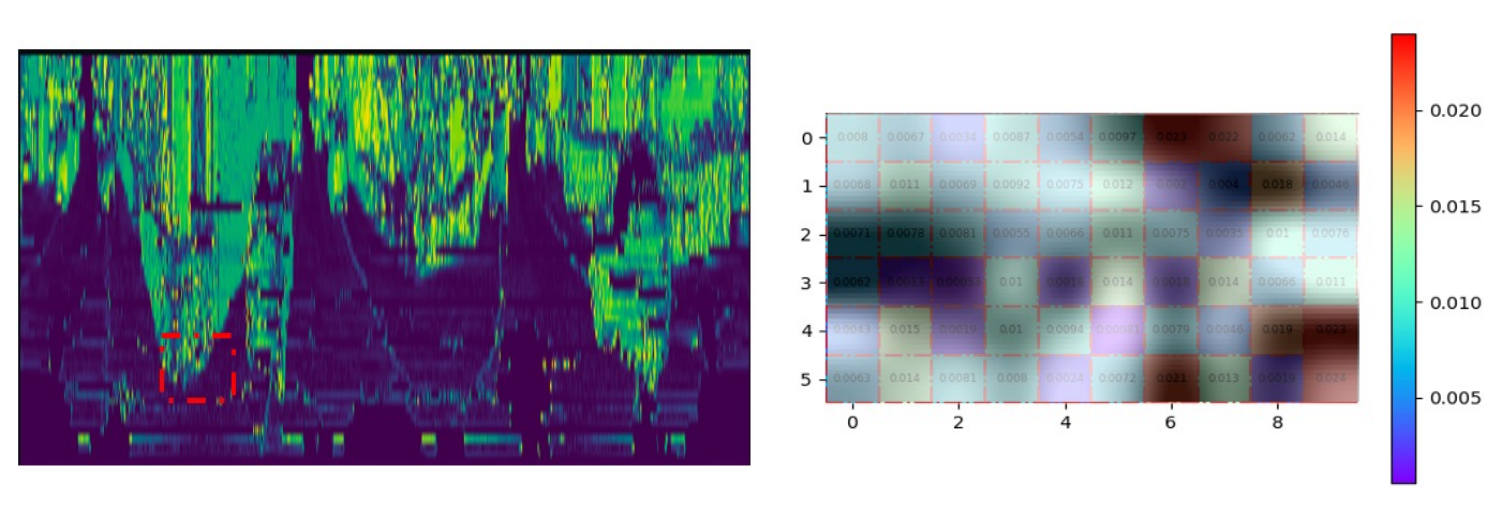}
        \label{overlap4}
    }
    
    \caption{Visualization of self-attention in T1 of head4 and cross-attention in T4 of head 4, red rectangle is query token, value of attended tokens is overlapped on source images.}
    \label{overlap}
\end{figure}

Additionally, as illustrated in Fig. \ref{attendedtoken}, we reshape the array of attended token values and overlap them onto the source modality image, as shown in Fig. \ref{overlap}. We present visualizations of head 4 in T1 for self-attention and T4 which has the most prominent cross-attention. In Fig. \ref{overlap0}, we select the first patch as the query token, and the highest value of the attended token is located in the same position as the query token or nearby positions. Moreover, the sum of attention values in the query's modality is around 0.75, indicating less attention in the other modality. Similar conditions exist in the LiDAR self-attention of T1.







In Fig. \ref{overlap1}, the query token is selected on the left side representing linear acceleration. The high values of the attended tokens are mainly located at the bottom of the road position. In Fig. \ref{overlap2}, the query token is located in a low-value position of the IMU signal image, and the values of the attended tokens are distributed almost equally. In comparison, in Fig. \ref{overlap3}, if the query token is selected in a high-value angular velocity position, we observe that the high weights are located on the main road and corners of the building. One possible reason for this is that the geometry shape of the corner contributes more when the car is turning with high angular velocity. To verify this hypothesis, we visualize the LiDAR-to-Inertial cross-attention in Fig. \ref{overlap4}, by selecting the corner patch as the query token, the high weights of attended tokens are mainly located in the right parts with high angular velocity value positions. 

Through the above visualization, it is shown that our proposed fusion strategy could promote or restrain some specific interactions between two modalities via assigning adaptive weights, which makes the whole incorporation more effective.

\subsection{Ablation study and overfitting problem}
\label{sec:ablation}

Since the proposed framework includes many modules, an exhaustive ablation study is necessary. We design the ablation study by separating the modules and evaluating their impact on performance, as shown in Table \ref{ablation}. The multi-layer strategy has been validated in MLF-VO \cite{mlfvo} and CE \cite{channelexchange}, and the multi-task regressor module is compared with the baseline 6D pose regressor in our previous work, CertainOdom \cite{certianodom}. Therefore, we keep these two modules as "open" status in all cases. 

The cases from (1) to (5) are designed to test three different fusion approaches: concatenate, SMAF, and Transformer. When selecting two fusion methods among these three as a combination such as (4) and (5), they actually include two opposite ways (e.g., concatenate between $v_{p}$ and $n_{p}$ and then SMAF with $i_{p}$ or inverse SMAF and concatenate). We only show the better results in Table \ref{ablation}. Case (7) is to verify the performance of the multi-scale module compared to the proposed TransFusionOdom (6).

The reason we not only list the testing results but also the training dataset results is that we have observed an overfitting problem in case (3). The issue of overfitting in learning-based odometry estimation tasks has rarely been mentioned. As we know, Transformer-based models are more data-hungry than CNN-based approaches \cite{transformerhungry}. Additionally, as mentioned before, the model size or complexity of Transformer-based models highly depends on the number and resolution of tokens \cite{vit}, and there is also a multi-layer fusion strategy inside the proposed framework, which increases the number of parameters. The bigger the model, the easier it is to overfit. Considering the above, it is necessary to check whether the proposed solutions and other Transformer-based fusion approaches are a good fit or overfitting model.

\begin{figure}[!t]

\subfigure[Loss curves in training and validation]{
\includegraphics[width=\columnwidth]{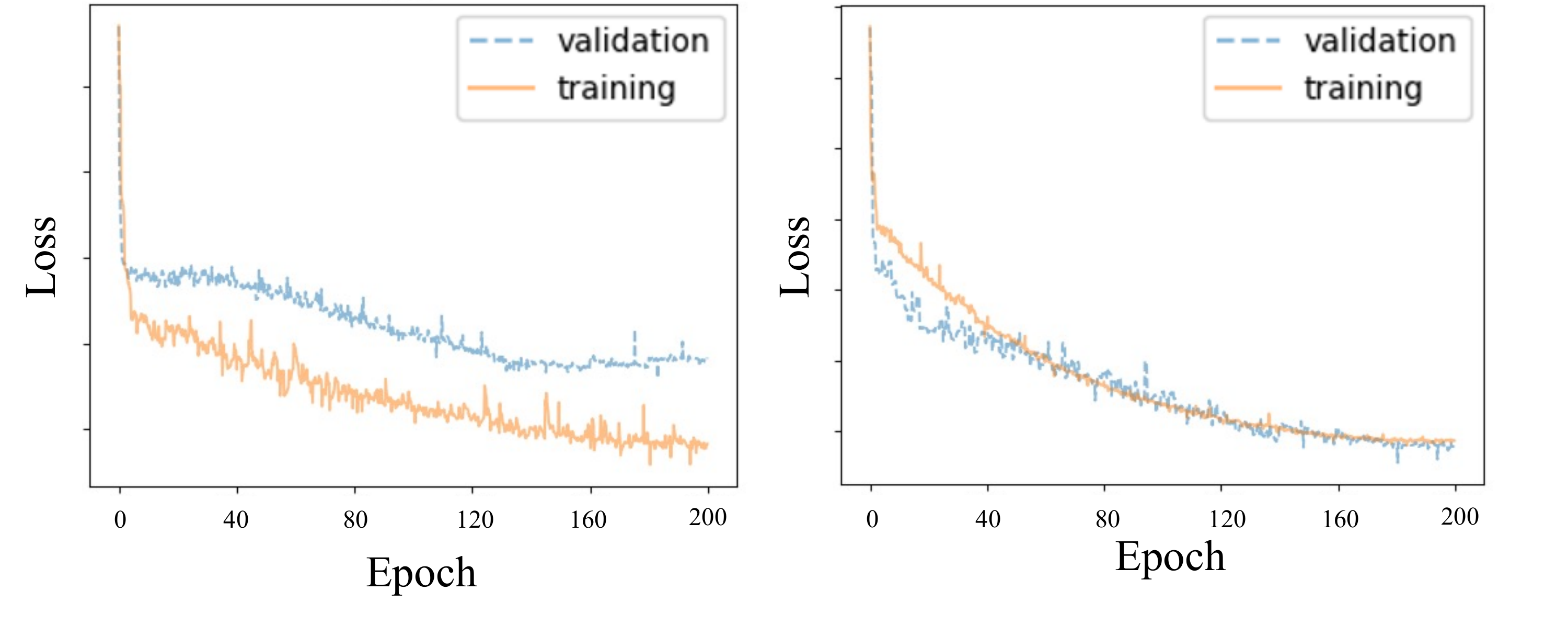}

}
\quad
\subfigure[Trajectory results of training data.]{

\includegraphics[width=8.5cm]{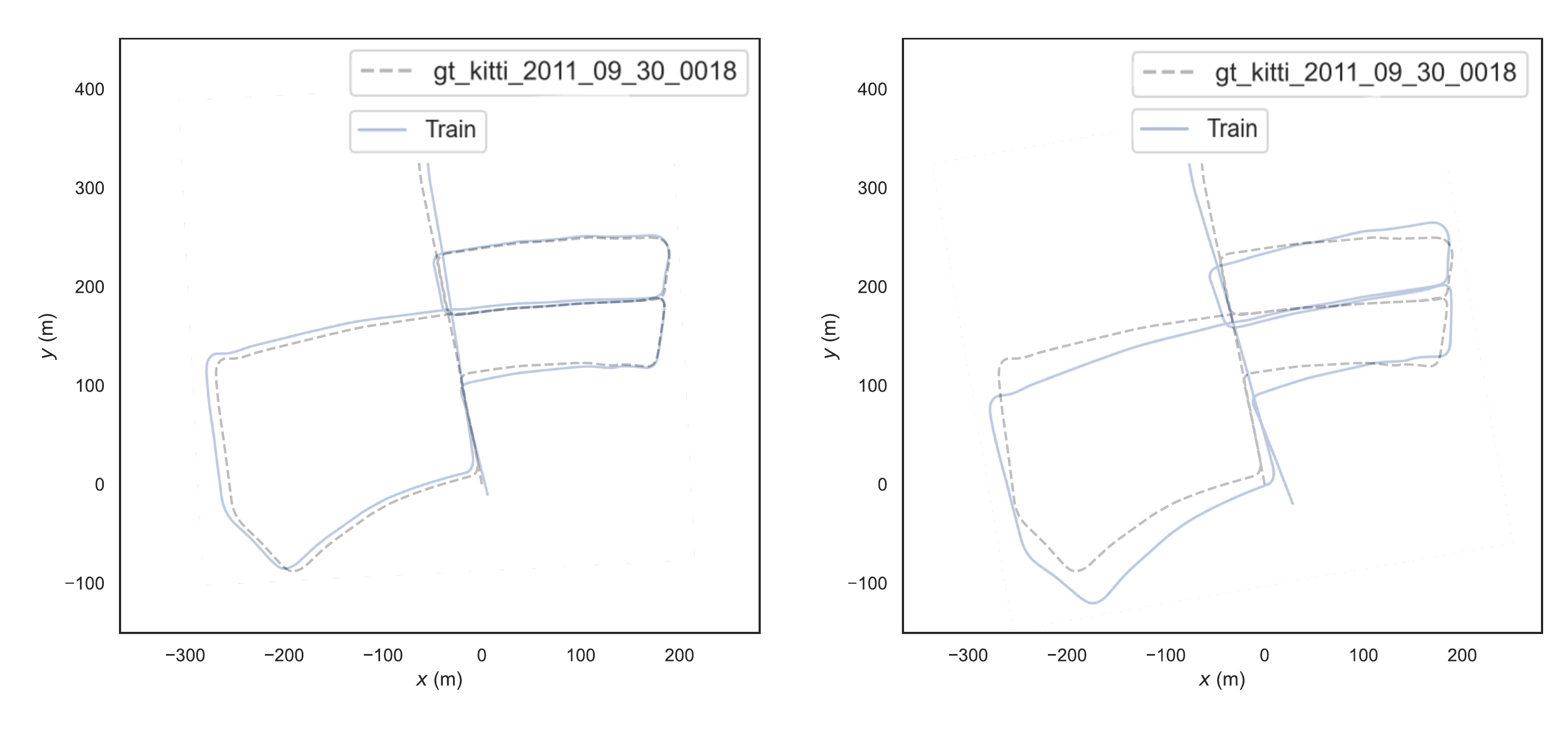}

}
\quad
\subfigure[Trajectory results of tested data.]{

\includegraphics[width=\columnwidth]{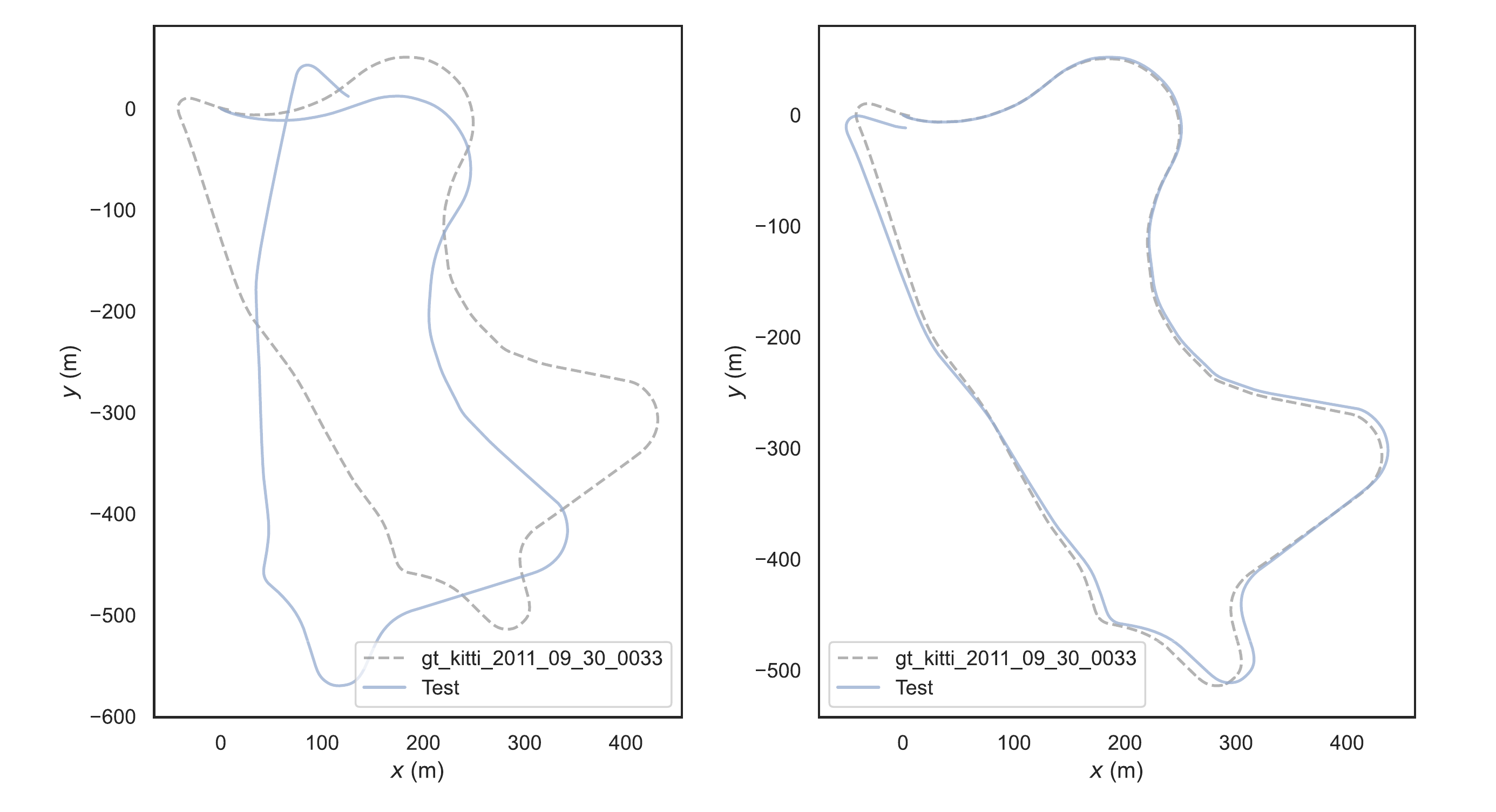}
}
\caption{Left: overfitting model, right: good-fitted model.}
\label{overfittingfig}
\end{figure}

In case (3), regardless of whether we use a Transformer to first fuse $v_{p}$ and $n_{p}$ as one general LiDAR modality and then deploy another Transformer to fuse it with $i_{p}$, or directly implement one Transformer with $v_{p}$, $n_{p}$ and $i_{p}$ as three modalities, the overfitting problem occurs. As shown in Fig. \ref{overfittingfig}, overfitting can be detected by the loss curves, also the overfitting model performs better than good-fitted model in training, but in testing, the result is worse than good-fitted model which is not what we expect. To overcome the overfitting problem, we implement a shared weights configuration between the backbones for vertex and normal features in Fig. \ref{architecture}. This way, the network can learn not only the common features but also its size can be reduced to avoid overfitting. Additionally, we could deploy data augmentation to increase the size of the training dataset, but it is outside the scope of this study.

From the statistical results in Table \ref{ablation}, it is clear how the performance of each fusion strategy compares. If we can overcome the overfitting problem with Transformer-based fusion, its performance is much better than that of SMAF and concatenation approaches. In case (7), we set a fixed resolution of input tokens from $l_{1}$ to $l_{4}$, which is equal to the $l_{4}$ resolution in TransfusionOdom (6). The reason for this is that the coarse-to-fine resolution is gradually reduced in multi-scale fusion. The higher the fixed resolution in (7), the more likely it is to lead to overfitting. Therefore, multi-scale strategy can also help avoid overfitting problems in Transformer-based fusion.

\begin{figure*}[!t]
\centering
\subfigure[Uncertainty estimation in translation and rotation, top row: 0035, bottom row: 0046.]{
\includegraphics[width=\textwidth]{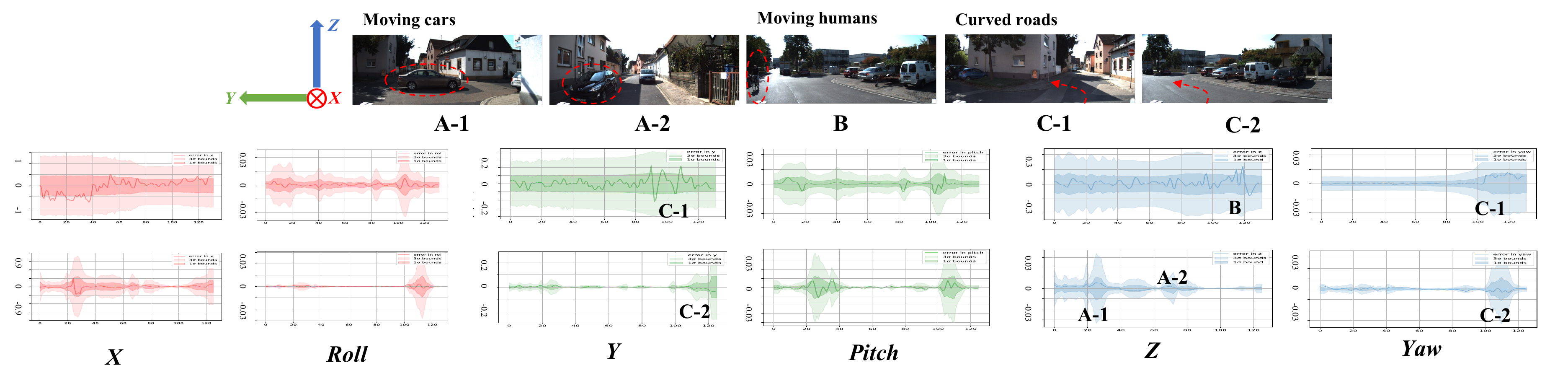}
\label{uncertainty}
}
\quad
\subfigure[Trajectory evaluation in translation and rotation with ground truth, left: 0035, right: 0046.]{
\includegraphics[width=\textwidth]{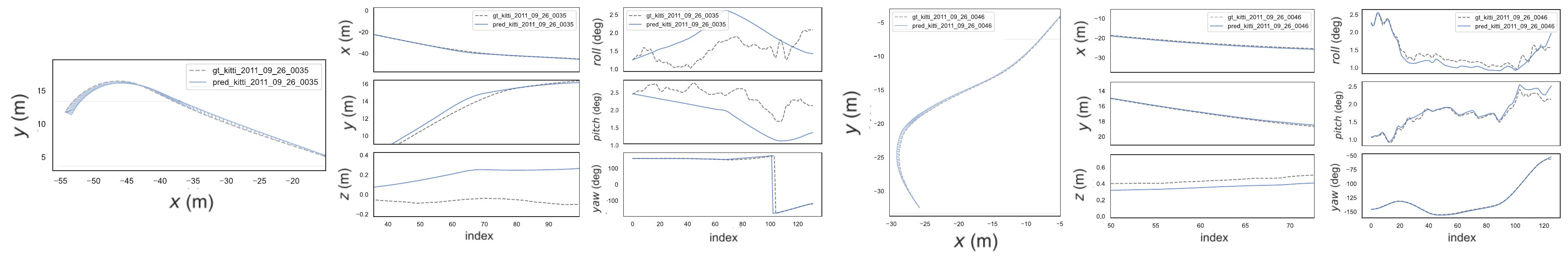}
\label{uncertaintytraj}

}

\caption{Uncertainty estimation evaluation and its corresponding trajectory results with ground truth on two KITTI sequences. }
\end{figure*}

\begin{table*}[!t]
\caption{Ablation study results}
\renewcommand\arraystretch{1.5}
\begin{center}
\scalebox{0.85}{
\begin{tabular}{ccccccccccc}
\hline
\multicolumn{1}{c|}{}                         & \multicolumn{1}{c|}{}                              & \multicolumn{1}{c|}{}                             & \multicolumn{1}{c|}{}                              & \multicolumn{1}{c|}{}                          & \multicolumn{1}{c|}{}                       & \multicolumn{1}{c|}{}                              & \multicolumn{2}{c|}{{ Average of KITTI Training}} & \multicolumn{2}{c}{{ Average of KITTI Testing}} \\ \cline{8-11} 
\multicolumn{1}{c|}{\multirow{-2}{*}{{ Module / Case}}} & \multicolumn{1}{c|}{\multirow{-2}{*}{{ Multi-layer}}} & \multicolumn{1}{c|}{\multirow{-2}{*}{{ Multi-task}}} & \multicolumn{1}{c|}{\multirow{-2}{*}{{ Multi-scale}}} & \multicolumn{1}{c|}{\multirow{-2}{*}{{ Concat.}}} & \multicolumn{1}{c|}{\multirow{-2}{*}{{ SMAF}}} & \multicolumn{1}{c|}{\multirow{-2}{*}{{ Transformer}}} & \multicolumn{1}{c|}{${t_{rel}}$(\%)}         & \multicolumn{1}{c|}{${r_{rel}}$ ($^{\circ}$)}        & \multicolumn{1}{c|}{${t_{rel}}$(\%)}       & \multicolumn{1}{c}{${r_{rel}}$ ($^{\circ}$)}       \\ \hline
\multicolumn{1}{c|}{{ (1)}}                      & { \checkmark}                          & { \checkmark}                         & { \checkmark}                          & { \checkmark}                      &                                             &                                                    &     10.82       &                8.32                   &     11.25     &    8.95                            \\ \cline{1-1}
\multicolumn{1}{c|}{{ (2)}}                      & { \checkmark}                          & { \checkmark}                         & { \checkmark}                          &                                                & { \checkmark}                   &                                                    &      8.24      &                6.12                   &      8.69    &    6.47                           \\  \cline{1-1}
\multicolumn{1}{c|}{{ (3)}}                      & { \checkmark}                          & { \checkmark}                         & { \checkmark}                          &                                                &                                           & { \checkmark}                          &       \textbf{0.49}     &                   \textbf{0.62}                &     3.52     &    4.66                            \\ \cline{1-1}
\multicolumn{1}{c|}{{ (4)}}                      & { \checkmark}                          & { \checkmark}                         & { \checkmark}                          & { \checkmark}                      & { \checkmark}                   &                                                    &       4.67     &   4.89                                &       4.72   &          4.93                      \\ \cline{1-1}
\multicolumn{1}{c|}{{ (5)}}                      & { \checkmark}                          & { \checkmark}                         & { \checkmark}                          & { \checkmark}                      &                                             & { \checkmark}                          &       3.53     &   3.72                                &    3.75      &           3.98                     \\ \cline{1-1}
\multicolumn{1}{c|}{{ (6) TransFusionOdom}}      & { \checkmark}                          & { \checkmark}                         & { \checkmark}                          &                                                & { \checkmark}                   & { \checkmark}                       &    0.52        & 0.67                                   &     0.61     &      0.71                          \\ \cline{1-1}
\multicolumn{1}{c|}{{ (7)}}                      & { \checkmark}                          & { \checkmark}                         &                                                    &                                                & { \checkmark}                   & { \checkmark}                          &   2.97         &      2.63                             &   3.12       &     2.88                           \\ \hline
\multicolumn{11}{l}{• ${t_{rel}}$: average sequence translational RMSE (\%) on the length of 100m, 200m, ..., 800m.}                                                                                                                                                            \\
\multicolumn{11}{l}{• ${r_{rel}}$: average sequence rotational RMSE ($^{\circ}$/100m) on the length of 100m, 200m, ..., 800m.}                                                                        
\end{tabular}
\label{ablation}
}
\end{center}
\end{table*}

\subsection{Positioning results on KITTI dataset}
\begin{figure}[t]
\centerline{\includegraphics[width=\columnwidth]{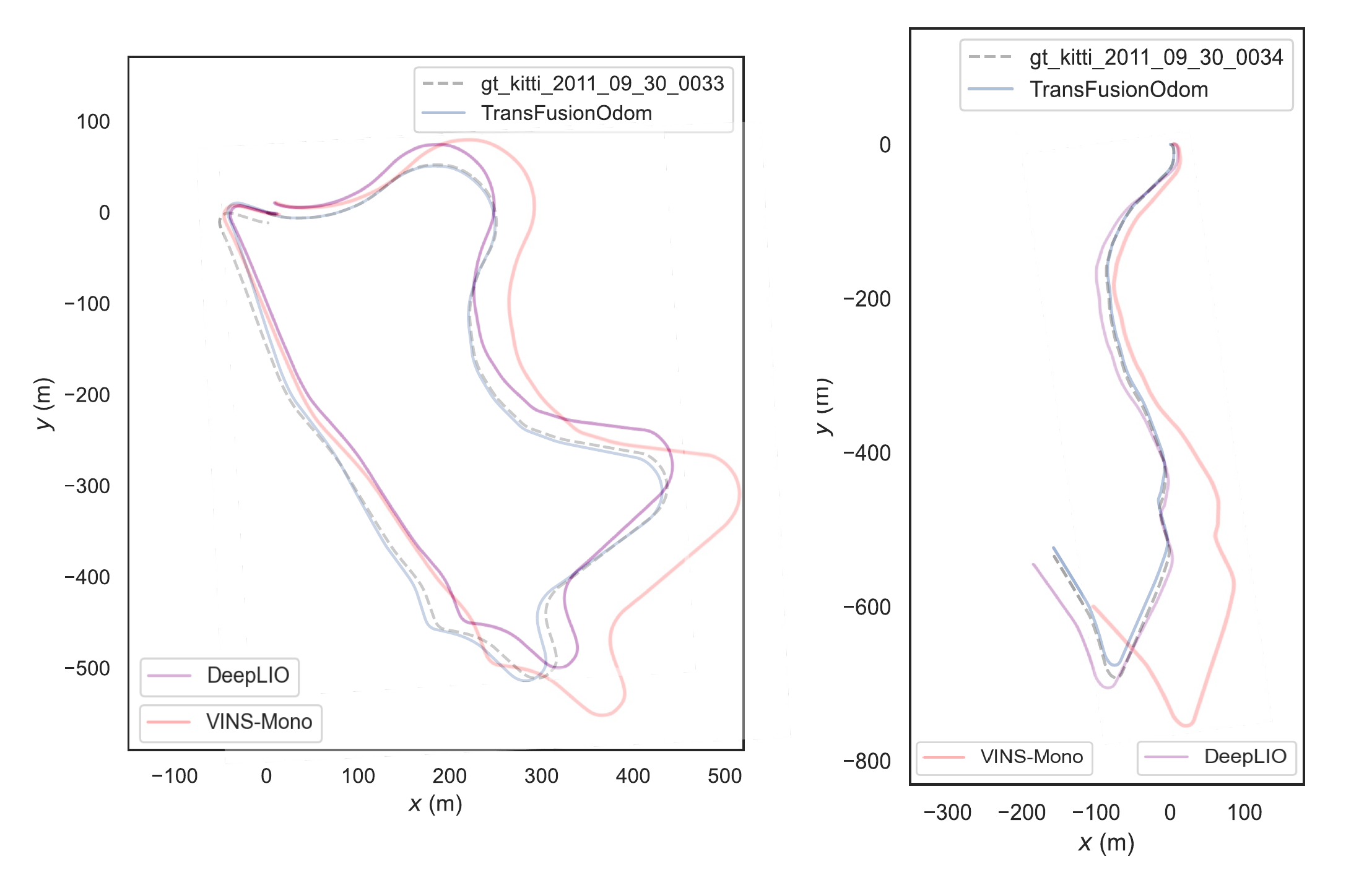}}
\caption{Trajectory evaluation in KITTI 09 and 10 sequence with related works.}
\label{trajectoryplot}
\end{figure}

We show the positioning results evaluated qualitatively in Fig. \ref{trajectoryplot} and quantitatively in Table \ref{positioning}, which compared with optimization-based VINS-Mono\cite{vins} and learning-based fusion approaches, DeepLIO\cite{deeplio}, EMA-VIO \cite{emavio}, LeGO-LOAM \cite{legoloam} and so on. The results show that our TransFusionOdom could achieve competitive performance among these related works. Compared with VINS-Mono \cite{vins}, the advantage of our proposed approach not only exists in fusion stage but also in front-end feature extraction which we mentioned in Section II.A. Besides, the improvement compared with EMA-VIO \cite{emavio} which also deploys Transformer-based approach for fusion, is possibly that the multi-layer fusion module aggregates the LiDAR and inertial data at different scale \cite{mlfvo} \cite{channelexchange}.

\begin{table*}[!t]
\caption{Positioning results on KITTI dataset}
\renewcommand\arraystretch{1.5}
\begin{center}
\scalebox{0.86}{
\begin{tabular}{ccccccccccccccc}
\hline
\multicolumn{1}{c|}{}                      & \multicolumn{2}{c|}{{ VINS-Mono \cite{emavio}}}                                             & \multicolumn{2}{c|}{{ DeepLIO \cite{deeplio}}}                                               & \multicolumn{2}{c|}{{ EMA-VIO \cite{emavio}}}                                               & \multicolumn{2}{c|}{{ CertainOdom \cite{certianodom}}}                                           & \multicolumn{2}{c|}{{ LeGO-LOAM \cite{litamin2}}}                                             & \multicolumn{2}{c|}{{ LO-Net \cite{lonet}}}                                                & \multicolumn{2}{c}{{ TransFusionOdom}}                             \\ \cline{2-15} 
\multicolumn{1}{c|}{\multirow{-2}{*}{{ Seq}}} & \multicolumn{1}{c|}{{ ${t_{rel}}$(\%)}} & \multicolumn{1}{c|}{{ ${r_{rel}}$ ($^{\circ}$)}} & \multicolumn{1}{c|}{{ ${t_{rel}}$(\%)}} & \multicolumn{1}{c|}{{ ${r_{rel}}$ ($^{\circ}$)}} & \multicolumn{1}{c|}{{ ${t_{rel}}$(\%)}} & \multicolumn{1}{c|}{{ ${r_{rel}}$ ($^{\circ}$)}} & \multicolumn{1}{c|}{{ ${t_{rel}}$(\%)}} & \multicolumn{1}{c|}{{ ${r_{rel}}$ ($^{\circ}$)}} & \multicolumn{1}{c|}{{ ${t_{rel}}$(\%)}} & \multicolumn{1}{c|}{{ ${r_{rel}}$ ($^{\circ}$)}} & \multicolumn{1}{c|}{{ tel}} & \multicolumn{1}{c|}{{ ${r_{rel}}$ ($^{\circ}$)}} & \multicolumn{1}{c|}{{ ${t_{rel}}$(\%)}} & { ${r_{rel}}$ ($^{\circ}$)}           \\ \hline
\multicolumn{1}{c|}{{ 09}}                    & { 41.4}                     & { 2.41}                     & { 4.4}                      & { 1.21}                     & { 8.86}                     & { 1.54}                     & { 0.53}                     & { 1.25}                     & { 0.98}                     & { 1.97}                     & { 1.37}                     & { \textbf{0.58}}                     & { \textbf{0.49}}            & { 0.63} \\ \cline{1-1}
\multicolumn{1}{c|}{{ 10}}                    & { 20.35}                    & { 2.73}                     & { 4.0}                      & { 1.51}                     & { 7.46}                     & { 2.26}                     & { 0.82}                     & { 1.06}                     & { 0.92}                     & { 2.21}                     & { 1.8}                     & { 0.93}                     & { \textbf{0.72}}            & { \textbf{0.78}} \\ \cline{1-1}
\multicolumn{1}{c|}{{ Avg.}}                  & { 30.91}                    & { 2.57}                     & { 4.2}                     & { 1.36}                     & { 8.07}                     & { 1.90}                     & { 0.68}                     & { 1.16}                     & { 0.95}                     & { 2.09}                     & { 1.59}                     & { 0.76}            & { \textbf{0.61}}            & { \textbf{0.71}}          \\ \hline
\multicolumn{15}{l}{• ${t_{rel}}$: average sequence translational RMSE (\%) on the length of 100m, 200m, ..., 800m.}                                                                                         \\
\multicolumn{15}{l}{• ${r_{rel}}$: average sequence rotational RMSE ($^{\circ}$/100m) on the length of 100m, 200m, ..., 800m.}                                                                                                                                                                                                   \\
\multicolumn{15}{l}{• The performance results of existed methods are collected from the cited literature.}                                                                                        
\end{tabular}
}
\end{center}
\label{positioning}
\end{table*}

\subsection{Uncertainty results on KITTI dataset}
As a byproduct of our proposed loss function, the uncertainty of predicted 6D pose could be used for down-streaming task, such as navigation, path planning and so on. In TransFusionOdom, the uncertainty regressed by multi-decoders is used to weigh the error between translation and rotation, the details we have discussed in our previous work CertainOdom \cite{certianodom}. The difference is that, through the experimental results, we implement Euler angles instead of quaternion as the representation of rotation in CertainOdom, which allows us to inference the uncertainty estimation in terms of roll, pitch, and yaw directions rather than a general rotation metric.

As shown in Fig. \ref{uncertainty}, we plot the error and its uncertainty of relative pose prediction. Most of the error has been bounded in the 1$\sigma$ area, and the fluctuation of uncertainty estimation also reflects the trend of error, which means the accuracy of the predicted uncertainty. Moreover, we observe that i) when encountering moving cars and humans, the uncertainty in the z-axis increases (as A-1/2 and B). ii) when the car is turning into the y-direction, it reflects in the uncertainty of the y-axis and yaw angle (as C-1/2). iii) The x direction has larger uncertainty than y and z axes, since the car travels the longest in x direction. The higher value of general uncertainty in the top row than the bottom row of Fig. \ref{uncertainty} could be validated in the trajectory comparison in Fig. \ref{uncertaintytraj}, where both the error in translation and rotation of the 0035 sequence are bigger than that of the 0046 sequence. Although the uncertainty estimation can be explained qualitatively and quantitatively in most moments, there are still some noisy values that cannot be interpreted. One possible reason is that this supervised uncertainty estimation approach is sensitive to out-of-distribution data \cite{due} \cite{ood} \cite{balancedvio} \cite{mdnvo} \cite{certianodom}.



\subsection{Gazebo-based synthetic multi-modal dataset}

\begin{figure}[t]
\centerline{\includegraphics[width=\columnwidth]{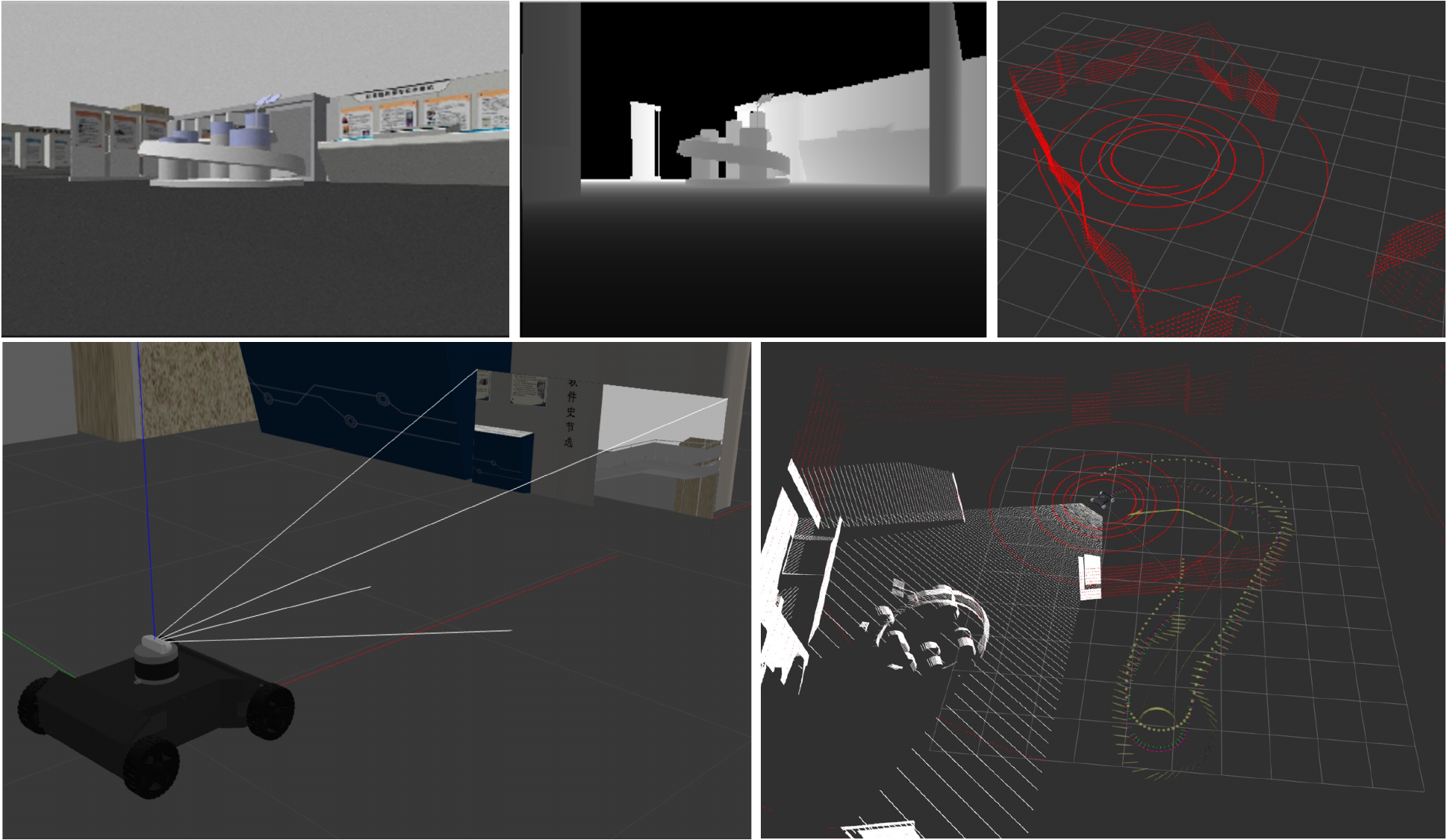}}
\caption{Publicly available synthetic multi-modal dataset for odometry estimation.}
\label{dataset}
\end{figure}

\begin{figure}[!t]
\centerline{\includegraphics[width=\columnwidth]{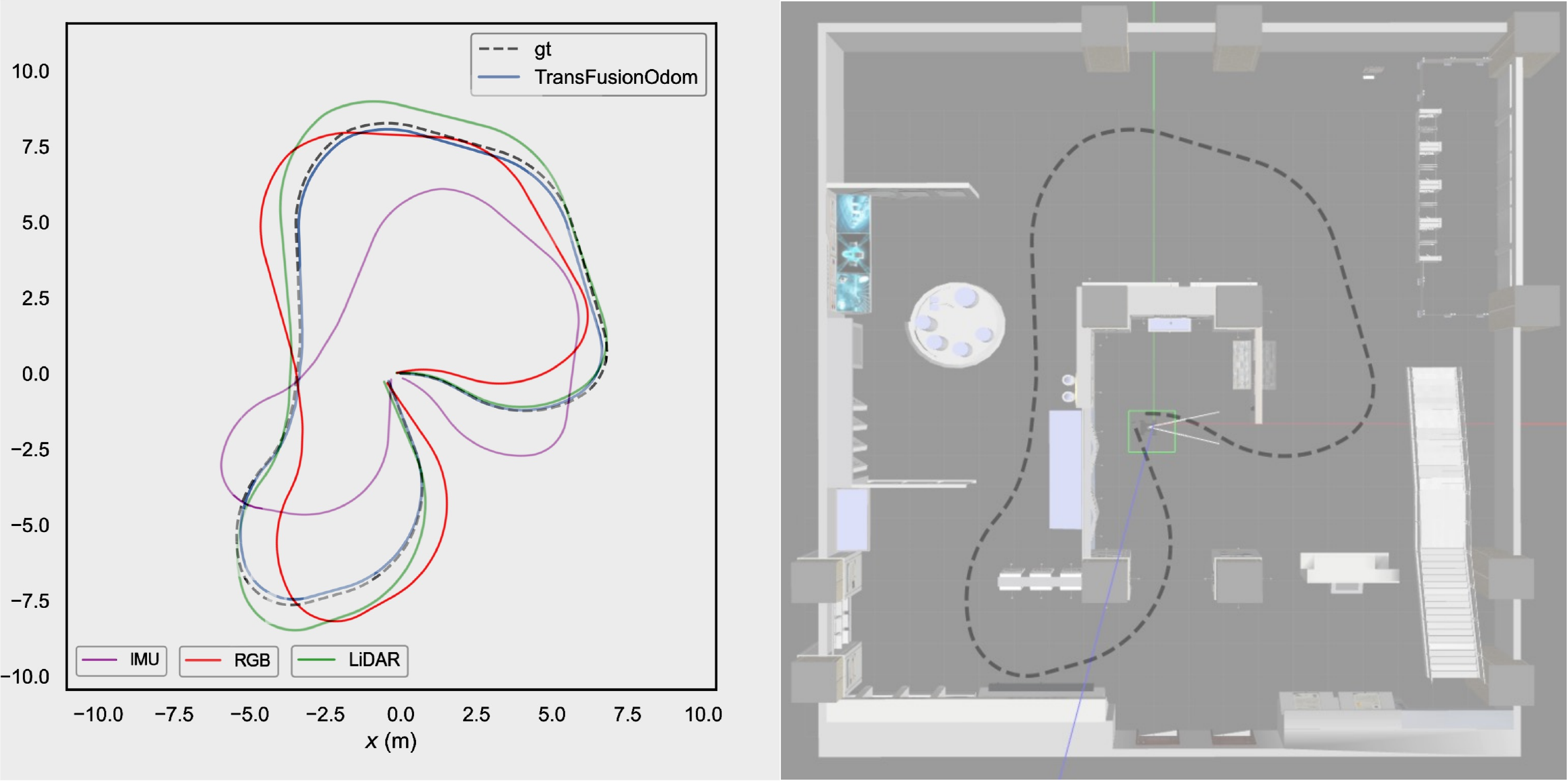}}
\caption{Comparison between uni-modal and TransFusionOdom on synthetic dataset.}
\label{unimodal}
\end{figure}

\begin{table}[!t]
\caption{Generalization experiments on synthetic multi-modal dataset}
\renewcommand\arraystretch{1.5}
\begin{center}
\scalebox{0.83}{
\begin{tabular}{c|cccccc}
\hline
{ Modality}    & \multicolumn{1}{c|}{{ LiDAR}} & \multicolumn{1}{c|}{{ RGB}} & \multicolumn{1}{c|}{{ Depth}} & \multicolumn{1}{c|}{{ IMU}} & \multicolumn{1}{c|}{{ \makecell[c]{Position \\ RMSE (m)}}} & { \makecell[c]{Attitude \\ RMSE ($^{\circ}$)} }\\ \hline
{ LiDAR-based} & { \checkmark}                         &                          &                            &                          &                    10.32      &   8.19  \\ \cline{1-1}
{ L-(2)}       & { \checkmark}                         & { \checkmark}                       &                            &                          &                      7.32    &  6.73   \\ \cline{1-1}
{ L-(3)}       & { \checkmark}                         &                          & { \checkmark}                         &                          &                      8.94    &  7.59   \\ \cline{1-1}
{ L-(4)}       & { \checkmark}                         &                          &                            & { \checkmark}                       &                     4.67   &  \textbf{2.83}   \\ \cline{1-1}
{ RGB-based}   &                            & { \checkmark}                       &                            &                          &                   13.47       &  10.76   \\ \cline{1-1}
{ R-(2)}       &                            & { \checkmark}                       & { \checkmark}                         &                          &                        9.32  & 8.17    \\ \cline{1-1}
{ R-(3)}       &                            & { \checkmark}                       &                            & { \checkmark}                       &           6.33               &  3.72   \\ \cline{1-1}
{ R-(4)}       &                            & { \checkmark}                       & { \checkmark}                         & { \checkmark}                       &                 \textbf{4.06}         &   2.99  \\ \hline
\end{tabular}
}
\end{center}
\label{generalization}
\end{table}

 We publish a synthetic multi-modal dataset for odometry estimation that was collected in the Gazebo simulator, as shown in Fig. \ref{dataset}. This dataset provides multi-sensor data, including Velodyne VLP-16 3D LiDAR, Realsense D435 RGB-D camera, IMU, and can also easily integrate other sensors. The ground truth is directly obtained from the simulation state. The dataset includes 5 scenarios/maps, each one is collected 5 times (3 for training, 2 for validation) multi-modal data using randomly trajectories generation. In the future, the Gazebo world maps can be replaced and even simulate moving objects. Using this dataset, the algorithms developed in this community can be conveniently tested.

We use this dataset to confirm the generalization ability of the proposed fusion strategy to other combinations of modalities. We separate the combinations into LiDAR-based and RGB-based, which are commonly configured for robotics perception and are listed in Table \ref{generalization}. We did not test the combinations with 4 or 3 modalities that include LiDAR because, as in TransFusionOdom, LiDAR includes vertex and normal sub-modalities. Adding more modalities can lead to overfitting problems. All modalities were converted to image type as input.

In Fig. \ref{unimodal}, we present a trajectory comparison between uni-modal and proposed TransFusionOdom. It turns out that it is difficult to obtain acceptable results by only using inertial data with Transformer architecture compared to use only vision-based modalities such as RGB and LiDAR. Moreover, unlike the common problem in geometry-based solutions, which is the scale ambiguity \cite{hvionet} \cite{viionet}, thanks to the f2f and f2g constraints, the scale problem is not so obvious in the learning-based approach. This similar conclusion was also observed in DeepLIO \cite{deeplio}.

In Table \ref{generalization}, there are different combinations of exteroceptive sensors (L-2, L-3, and R-2) and proprioceptive with exteroceptive sensors (L-4, R-3, and R-4). The latter one has a better performance than the former combination, which is the same as our common sense. Besides, we observe that IMU contributes more to attitude than position. In L-4 and R-3, the performance has on average increased by 53$\%$ in position and 65$\%$ in attitude, respectively. In R-4, RGB and depth are homogeneous modalities fused by SMAF and then integrated with IMU data using Transformer, which obtains the best performance in position. Generally, these experiments are not designed to test which combination is the best, but to verify that our fusion strategy can generalize to different modalities to achieve better performance than uni-modal instead of only LiDAR with IMU in TransFusionOdom.

\section{Conclusion and future work}
In this study, we present TransFusionOdom, a Transformer-based supervised end-to-end LiDAR-inertial odometry framework. We mainly discuss the performance of different fusion strategies that involve the homogeneous and heterogeneous modalities. We demonstrate a general approach to visualize self and cross attention inside TransFusionOdom, which enables us to interpret how different modalities interact with each other via attention mechanisms. We conduct an exhaustive ablation study to check the performance of each module in the framework. Additionally, we illustrate and discuss the overfitting problem in odometry estimation caused by Transformer-based fusion strategy. We also collect and made publicly available a synthetic dataset to validate the well generalization ability of the proposed fusion strategy on different modalities. The competitive odometry and uncertainty estimation results are evaluated qualitatively and quantitatively on KITTI dataset.

Back to the previous question: \textit{How should we perform fusion among different modalities in a supervised sensor fusion odometry estimation task?}. In order to achieve the adaptive incorporation between different modalities, we propose the multi-attention fusion module. While the advanced Transformer-based architecture could boost the effective internal interactions, blindly increasing the size of the model can lead to several issues. Given this consideration, we deploy the lightweight MLP-based SMAF to fuse the homogeneous modalities. The Transformer-based fusion is conducted in the heterogeneous modalities, which is more complex and challenging \cite{coupled}. Combined with multi-scale and layer fusion module, a generic and flexible fusion strategy is developed and validated in the study.

However, vision Transformer suffers from high redundancy by only focusing on local features or self-attention domains in shallow layers \cite{supertoken}, as we discussed in the visualization of the attention matrix in the early fusion stage. If we can effectively achieve global context modeling at the early stage of the Transformer-based architecture, we can make the neural network model lightweight, which is beneficial for easy training and real-world applications.

\begin{table}[t]
\caption{Network architecture of implementation.}
\renewcommand\arraystretch{1.6}
\scalebox{0.85}{
\begin{tabular}{l|ll}
\hline
{[}input{]}                                                                         & \multicolumn{2}{l}{\begin{tabular}[c]{@{}l@{}}Two stacked LiDAR point cloud and\\ inertial measurement between them.\end{tabular}}                                                                                                                                                                                                                  \\ \hline
\multirow{2}{*}{{[}Pre-processing{]}}                                               & \multicolumn{2}{l}{\begin{tabular}[c]{@{}l@{}}Two stacked vertex and normal maps: \\ Batch size (B)*Window size(W)*720*60*6\end{tabular}}                                                                                                    \\ \cline{2-3} 
                                                      & \multicolumn{2}{l}{Inertial digital image: B*W*10*6}      \\ \hline
{[}SMAF in Layer 1{]}                                                               & \multicolumn{2}{l}{\begin{tabular}[c]{@{}l@{}}Vertex/normal map --\textgreater Conv2d(kernel\_size=$7^{2}$, stride=$2^{2}$,\\  padding=3)--\textgreater BatchNorm2d--\textgreater{}ReLU\\                    --\textgreater MaxPool2d(kernel\_size=$3^{2}$, stride=$2^{2}$, padding=1)\\                    --\textgreater ResNet34.layer1(64,3,2e-2); output=$[v_{l1};n_{l1}]$\\ $SMAF_{l1}$= $FC_{v1}$[$v_{l1}$;$n_{l1}$]*$v_{l1}$ + $FC_{n1}$[$v_{l1}$;$n_{l1}$]*$n_{l1}$ \end{tabular}}                                                                                                                                                                                                                                         \\ \hline
\begin{tabular}[c]{@{}l@{}}{[}Transformer-based \\ in Layer1{]}\end{tabular}        & \multicolumn{2}{l}{\begin{tabular}[c]{@{}l@{}}Inertial digital image --\textgreater Conv2d(kernel\_size=$7^{2}$, stride=$2^{2}$, \\ padding=3) --\textgreater BatchNorm2d--\textgreater{}ReLU\\                     --\textgreater MaxPool2d(kernel\_size=3\textasciicircum{}2, stride=2\textasciicircum{}2, padding=1)\\                     --\textgreater ResNet18.layer1(64,2,2e-2); output=$I_{l1}$\\ $I_{pool1}$/$SMAF_{pool1}$ = AdaptiveAvgPool2d(15,18)\\ $I_{T1}$,$SMAF_{T1}$=$Transformer_{l1}$[$I_{pool1}$, $SMAF_{pool1}$]\\ $SMAF_{T1}$ --\textgreater interpolate($SMAF_{l1}$.size(), mode='bilinear')\\ $I_{T1}$--\textgreater interpolate($I_{l1}$.size(), mode='bilinear')\\ $S_{l1}$ = $SMAF_{l1}$+$SMAF_{T1}$; $IMU_{l1}$ = $I_{l1}$+$I_{T1}$\end{tabular}} \\ \hline
[SMAF in Layer 2/3/4]                                                                 & \multicolumn{2}{l}{\begin{tabular}[c]{@{}l@{}} [$v_{l1/2/3}$+$S_{l1/2/3}$, $n_{l1/2/3}$+$S_{l1/2/3}$] \\ --\textgreater ResNet34.layer2/3/4(128/256/512,4/6/3,2e-2)\\ output = $[v_{l2/3/4};n_{l2/3/4}]$\\ $SMAF_{l2/3/4}$= $FC_{v2/3/4}$[$v_{l2/3/4}$;$n_{l2/3/4}$]*$v_{l2/3/4}$ + \\ $FC_{n2/3/4}$[$v_{l2/3/4}$;$n_{l2/3/4}$]*$n_{l2/3/4}$\end{tabular}}                                                                                                                                                           \\ \hline
\begin{tabular}[c]{@{}l@{}}{[}Transformer-based \\ in Layer2/3/4{]}\end{tabular}    & \multicolumn{2}{l}{\begin{tabular}[c]{@{}l@{}}$IMU_{l1/2/3}$  --\textgreater{}ResNet18.layer2/3/4(128/256/512,2/2/2,2e-2)\\ $I_{pool2/3/4}$/$SMAF_{pool2/3/4}$ = \\ AdaptiveAvgPool2d((12,15),(10,12),(6,10))\\$I_{T2/3/4}$,$SMAF_{T2/3/4}$= \\$Transformer_{l2/3/4}$[$I_{pool2/3/4}$, $SMAF_{pool2-4}$]\\$S_{l2/3/4}$ = $SMAF_{l2/3/4}$ + $SMAF_{T2/3/4}$\\ $IMU_{l2/3/4}$ = $I_{l2/3/4}$+$I_{T2/3/4}$\\ $output_{T4}$ = [$SMAF_{l4}$; $IMU_{l4}$]\end{tabular}}                                                                                                                \\ \hline
\begin{tabular}[c]{@{}l@{}}{[}6D pose and \\ uncertainty regressor{]}\end{tabular} & \multicolumn{2}{l}{\begin{tabular}[c]{@{}l@{}}4 different MLPs($output_{T4}$) for 4 decoders: \\ 2 Linear(1024,3) for translation and rotation + \\2 Linear(1024,3) for uncertainty in position and orientation \end{tabular}}                                                                                                                                \\ \hline
\end{tabular}
\label{tab2}}
\end{table}

\bibliographystyle{IEEEtran}   

\bibliography{IEEE_Sensors_J}

\end{document}